\documentclass{article}
\usepackage{makecell}
\usepackage{floatrow}
\usepackage{float}
\usepackage{PRIMEarxiv}
\usepackage{graphicx}
\usepackage[utf8]{inputenc} % allow utf-8 input
\usepackage[T1]{fontenc}    % use 8-bit T1 fonts
\usepackage{hyperref}       % hyperlinks
\usepackage{url}            % simple URL typesetting
\usepackage{booktabs}       % professional-quality tables
\usepackage{amsfonts}       % blackboard math symbols
\usepackage{nicefrac}       % compact symbols for 1/2, etc.
\usepackage{microtype}      % microtypography
\usepackage{lipsum}
\usepackage{fancyhdr}       % header
\usepackage{graphicx}       % graphics
\graphicspath{{media/}}     % organize your images and other figures under media/ folder
\usepackage{amsmath} 
\usepackage{threeparttable}
\usepackage{array}
\usepackage{footmisc,hyperref}
\usepackage{longtable}
\usepackage{subcaption}
\usepackage{multirow}
\usepackage[table]{xcolor}
\usepackage{multicol}
\usepackage{color}
\usepackage{hyperref}
\usepackage{appendix}
\usepackage{adjustbox}
\hypersetup{
colorlinks=true,
linkcolor=red,
filecolor=cyan,
urlcolor=cyan,
citecolor=cyan,
}
\definecolor{purple}{RGB}{204, 204, 255} % 自定义浅紫色
\definecolor{gray}{RGB}{229, 229, 229}
\usepackage{authblk}

\title{Biomed-DPT: Dual Modality Prompt Tuning for Biomedical Vision-Language Models}

\author[1]{Wei Peng}
\author[1]{Kang Liu}
\author[1,2]{Jianchen Hu*}
\author[1,2]{Meng Zhang}

\affil[1]{School of Future Technology, Xi'an Jiaotong University, Xi'an, China}
\affil[2]{School of Automation Science and Engineering, Xi'an Jiaotong University, Xi'an, China}

\begin{document}
\maketitle

%\footnotetext[1]{horace89@xjtu.edu.cn}

\begin{abstract}
Prompt learning is one of the most effective paradigms for adapting pre-trained vision-language models (VLMs) to the biomedical image classification tasks in few shot scenarios. However, most of the current prompt learning methods only used the text prompts and ignored the particular structures (such as the complex anatomical structures and subtle pathological features) in the biomedical images. In this work, we propose Biomed-DPT, a knowledge-enhanced dual modality prompt tuning technique. In designing the text prompt, Biomed-DPT constructs a dual prompt including the template-driven clinical prompts and the large language model (LLM)-driven domain-adapted prompts, then extracts the clinical knowledge from the domain-adapted prompts through the knowledge distillation technique. In designing the vision prompt, Biomed-DPT introduces the zero vector as a soft prompt to leverage attention re-weighting so that the focus on non-diagnostic regions and the recognition of non-critical pathological features are avoided. Biomed-DPT achieves an average classification accuracy of 66.14\% across 11 biomedical image datasets covering 9 modalities and 10 organs, with performance reaching 78.06\% in base classes and 75.97\% in novel classes, surpassing the Context Optimization (CoOp) method by 6.20\%, 3.78\%, and 8.04\%, respectively. Our code are available at \underline{https://github.com/Kanyooo/Biomed-DPT}.
\end{abstract}

\section{Introduction}
\label{sec:introduction}
\begin{figure*}[!t]
\centering
\begin{subfigure}{0.32\linewidth}
    \centering
    \includegraphics[width=0.95\linewidth]{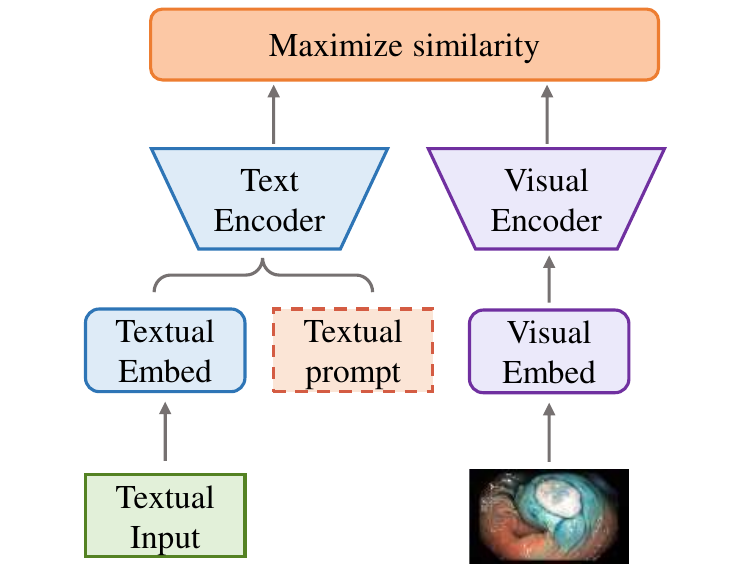}
    \caption{Textual-modal fine-tuning methods}
    \label{DPT_1}%文中引用该图片代号
\end{subfigure}
\centering
\begin{subfigure}{0.32\linewidth}
    \centering
    \includegraphics[width=0.95\linewidth]{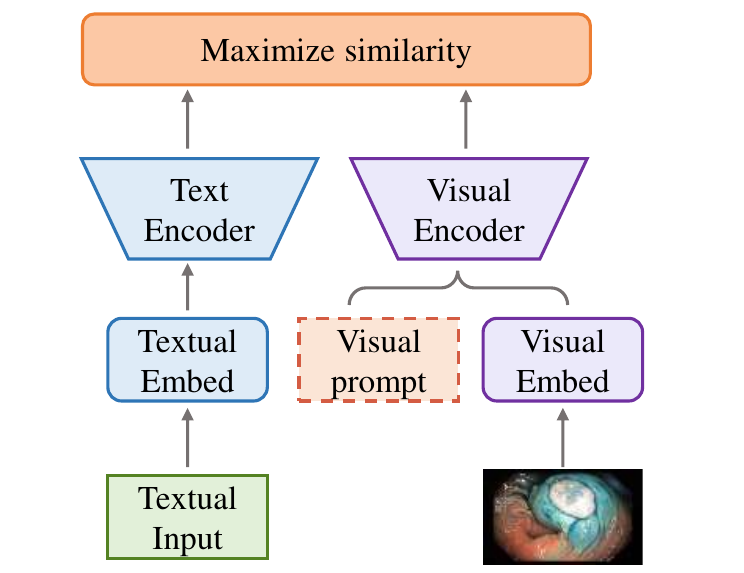}
    \caption{Visual-modal fine-tuning methods}
    \label{DPT_2}%文中引用该图片代号
\end{subfigure}
\centering
\begin{subfigure}{0.32\linewidth}
    \centering
    \includegraphics[width=0.95\linewidth]{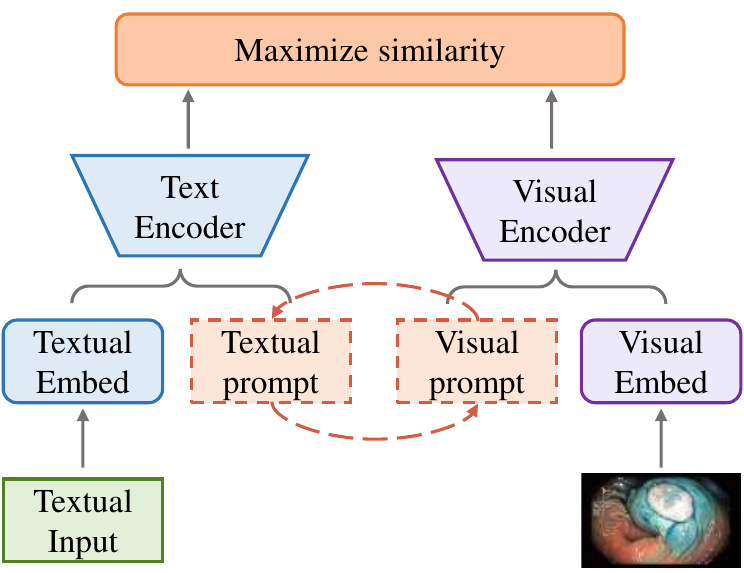}
    \caption{Dual-modal fine-tuning methods}
    \label{DPT_3}%文中引用该图片代号
\end{subfigure}
    \caption{Three categories of fine-tuning methods in VLMs. The dashed lines indicate potential interaction relationships between prompts.}
\end{figure*}

In recent years, pre-trained vision-language models (VLMs) such as CLIP~\cite{clip} and ALIGN~\cite{align} have made significant progress in zero-shot classification tasks for natural images. However, due to the high dimensionality and complexity of biomedical images, the development of large-scale VLMs in the biomedical field is relatively backward~\cite{LI2023126720}. At present, most pre-trained VLMs are trained on the natural image datasets, but features learned from natural image are significantly different from features for biomedical images. Additionally, the text prompts designed in the natural domain are usually naive, such as ``\texttt{a photo of a [CLASS].}'' which is completely different from the dense professional terminology and precise semantics of biomedical clinical reports. This dual-domain discrepancy between vision and text leads to poor performance of general models in biomedical image classification tasks~\cite{9412444}. In order to address this domain adaptation issue, many VLMs specifically designed for the biomedical field have been proposed (e.g., PMC-CLIP~\cite{lin2023pmc}, PubMedCLIP~\cite{pubmedclip}, BiomedCLIP~\cite{biomedclip}). These VLMs learned the association between biomedical images and clinical reports by pre-training on large-scale biomedical image-text pairs. In order to improve the adaptability in downstream tasks, transfer learning or fine-tuning of pre-trained VLMs are necessary. In order to this end, various Parameter-Efficient Fine-Tuning (PEFT) methods have been proposed (e.g., CoOp~\cite{coop}, Adapter~\cite{clipadapter}, and Linear Probe~\cite{clip}), which effectively enhance the performance on downstream image classification tasks by introducing only a fraction of learnable parameters. Based on the location of introduced parameters, fine-tuning methods can be categorized into the following three types: 1) Textual-modal fine-tuning methods (Figure~\ref{DPT_1}); 2) Visual-modal fine-tuning methods (Figure~\ref{DPT_2}); 3) Dual-modal fine-tuning methods (Figure~\ref{DPT_3}).

From a theoretical perspective, dual-modal fine-tuning methods, which integrate both visual and textual prompts, should theoretically outperform single-modal fine-tuning approaches (visual-only or textual-only), and single-modal fine-tuning methods should also surpass zero-shot learning benchmarks. However, the experimental results, as shown in Table~\ref{tab:avg_modal} (taking prompt learning as an example), reveal a significant reversal of this expected trend in the biomedical image domain as follows: 1) The accuracy of the visual-only modal (Visual-modal) is notably lower than the Zero-shot BiomedCLIP benchmark. 2) Due to the poor performance of visual prompts, dual-modal fine-tuning methods without visual-textual interaction underperform the Zero-shot BiomedCLIP benchmark in medical image classification tasks. 3) Although methods incorporating visual-textual modal interaction mechanisms show improvement over non-interactive approaches, their overall performance still falls short of textual-only prompt methods due to the inherent weakness of visual prompts. Therefore, designing effective visual prompts and textual prompts tailored for biomedical images is crucial.

\begin{table}[htbp]
\centering
\caption{The average classification accuracy (\%) obtained from 5 benchmarks, where s indicates the introduction of only one learnable parameter layer. (w) denotes with interaction, and (w/o) denotes without interaction.}
\label{tab:avg_modal}
\begin{tabular}{p{60pt}ccccc}
\toprule
Method & $K=1$ & $K=2$ & $K=4$ & $K=8$ & $K=16$ \\ 
\midrule
\rowcolor{gray}
\multicolumn{6}{c}{\textbf{Zero-shot Methods}} \\
BiomedCLIP~\cite{biomedclip} & \multicolumn{5}{c}{42.05}  \\
\midrule
\rowcolor{gray}
\multicolumn{6}{c}{\textbf{Textual-modal prompt learning}} \\
CoOp~\cite{coop} & 50.18 & 54.17 & 59.77 & 65.85 & 69.72  \\
\midrule
\rowcolor{gray}
\multicolumn{6}{c}{\textbf{Visual-modal prompt learning}} \\
VPT-s~\cite{vpt} &  21.67 & 22.65 & 20.99 & 26.08 & 24.31  \\
\midrule
\rowcolor{gray}
\multicolumn{6}{c}{\textbf{Dual-modal prompt learning}} \\
\makecell[l]{CoOp+VPT\\-s(w/o)~\cite{coop, vpt}} &  27.86 & 28.2  & 28.55 & 30.61 & 38.01   \\
Maple-s(w) ~\cite{maple} & 39.03 & 38.38 & 41.69 & 49.88    \\
\rowcolor{purple}
\makecell[l]{Biomed-DPT\\-s(w/o)(Ours)} & \textbf{59.03} & \textbf{61.27} & \textbf{66.12} & \textbf{70.76} & \textbf{73.51} \\
\bottomrule
\end{tabular}
\end{table}

\begin{figure}[!t]
    \centering
    \setlength{\tabcolsep}{1pt} % 设置列间距为1pt
    \renewcommand{\arraystretch}{0} 
    \setlength{\extrarowheight}{1pt} % 添加额外的行间距
    \begin{tabular}{c*{3}{c}} 
        % --- 第1行 ---
        \raisebox{1.3\height}{\rotatebox[origin=c]{90}{\footnotesize\textbf{Natural Image}}}\hspace{4pt} &
        \includegraphics[width=0.15\textwidth,height=0.10\textheight]{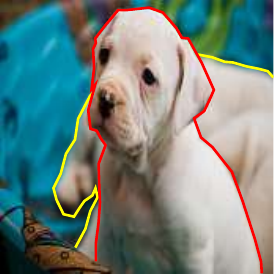} &
        \includegraphics[width=0.15\textwidth,height=0.10\textheight]{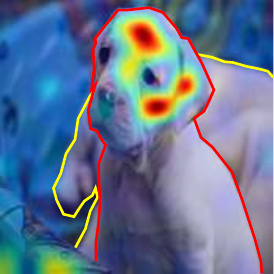} &
        \includegraphics[width=0.15\textwidth,height=0.10\textheight]{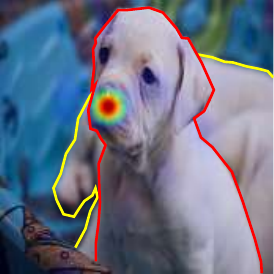} \\[3pt] % 添加行间距

        % --- 第2行 ---
        \raisebox{1.0\height}{\rotatebox[origin=c]{90}{\footnotesize\textbf{Biomedical Image}}}\hspace{4pt} &
        \includegraphics[width=0.15\textwidth,height=0.10\textheight]{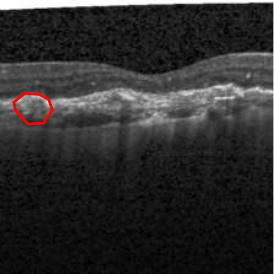} &
        \includegraphics[width=0.15\textwidth,height=0.10\textheight]{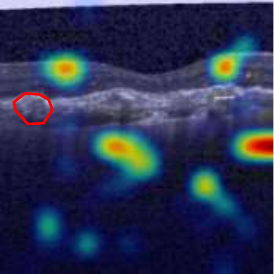} &
        \includegraphics[width=0.15\textwidth,height=0.10\textheight]{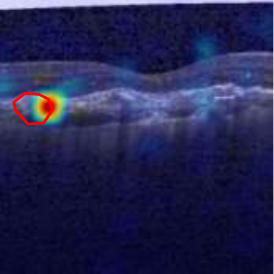} \\[5pt] % 添加行间距

        % --- 列标题（最后一行下方）---
        & \small (a) Image & \small (b) CLIP & \small (c) BiomedCLIP  \\
    \end{tabular}
    \caption{Effect of CLIP pre-tained on natural datasets vs BiomedCLIP pre-tained on biomedical datasets on visual saliency maps. Red lines indicate foreground regions, yellow lines represent background regions associated with object.}
    \label{fig:comparison}
\end{figure}

In visual representation, the biomedical image is fundamentally differ from natural optical image, as their generation process primarily relies on specific physical imaging principles~\cite{9363915}. These unique imaging mechanisms endow biomedical images with distinct visual patterns in grayscale distribution, texture features, and structural representation compared to the natural images~\cite{824822}. By comparing CLIP ~\cite{clip} with BiomedCLIP~\cite{biomedclip}, we observe that the natural image-trained VLM and biomedical image-trained VLM exhibit remarkly different patterns in sense of the attention distribution. As shown in Figure~\ref{fig:comparison}(b), CLIP ~\cite{clip} pre-trained on natural images displays global attention distributions, where the attention heatmap covers the extensive image regions and it shows deep attention in background areas. In contrast, as illustrated in Figure~\ref{fig:comparison}(c), BiomedCLIP~\cite{biomedclip} pre-trained on biomedical images demonstrates highly localized attention patterns, with the heatmap focused on specific anatomical or pathological regions (e.g., tumor size, morphology), while attention in background areas is significantly weaken. This discrepancy stems from the inherent differences between the two types of images: As depicted in Figure~\ref{fig:comparison}(a), the natural images typically exhibit clear foreground-background structures, prompting VLMs to focus on both subject and background, whereas the biomedical images present a binary division between diagnostically relevant and non-relevant regions.

More importantly, compared to the natural image, the biomedical image is more difficult to classify since the recognition of the pathological features requires more high-level information. Take the CheXpert dataset~\cite{chexpert} as an example, chest X-ray diagnosis requires the simultaneous identification of 14 types of pathological signs (e.g., cardiomegaly, lung opacity), most of which present as subtle density changes (e.g., abnormal lung field lucency) or minor displacements of anatomical structures. This high sensitivity to local features requires that VLMs possess extremely strong fine-grained feature detection capability, which makes the interpretation of biomedical images more challenging compared to natural images. Therefore, in biomedical image diagnosis, an ideal model should accurately focus on the critical areas, while effectively eliminate the distraction of attention to non-diagnostic areas. 

\begin{figure*}[htbp]
	\centering
	\begin{subfigure}{0.49\linewidth}
		\centering
		\includegraphics[width=0.98\linewidth]{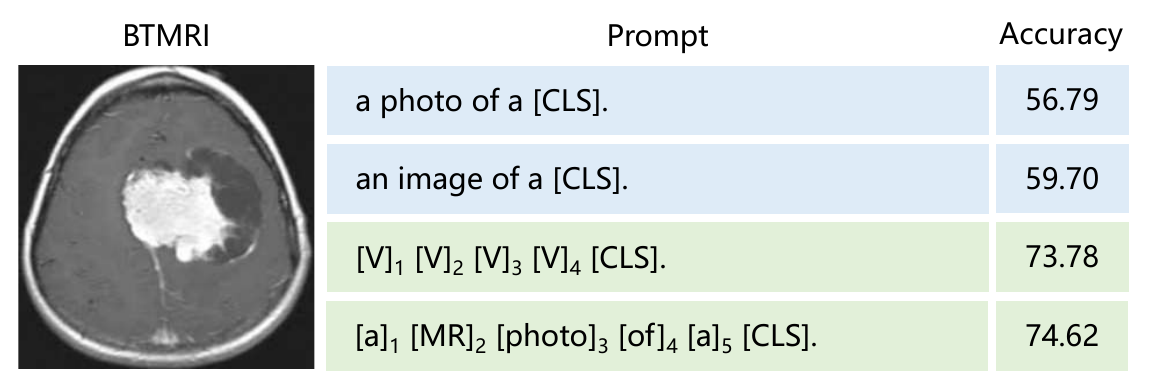}
		\caption{}
		\label{prompt_en_1}%文中引用该图片代号
	\end{subfigure}
	\centering
	\begin{subfigure}{0.49\linewidth}
		\centering
		\includegraphics[width=0.98\linewidth]{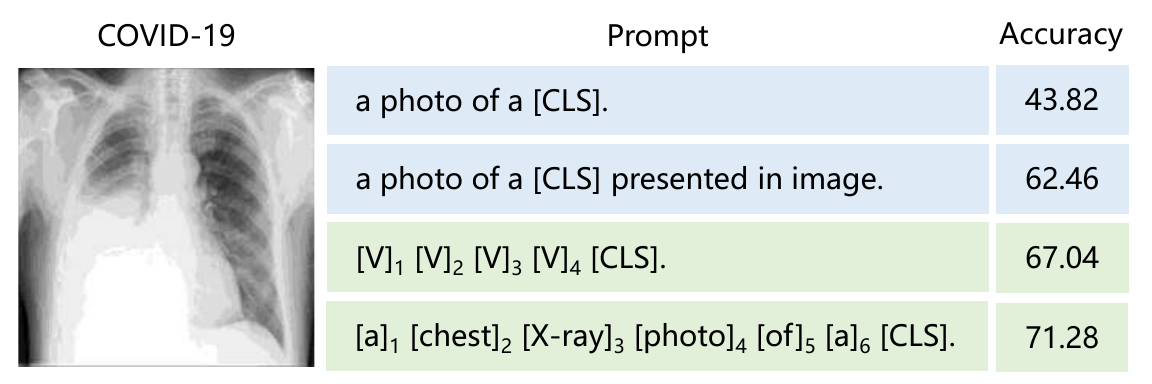}
		\caption{}
		\label{prompt_en_2}%文中引用该图片代号
	\end{subfigure}
	\centering
	\begin{subfigure}{0.49\linewidth}
		\centering
		\includegraphics[width=0.98\linewidth]{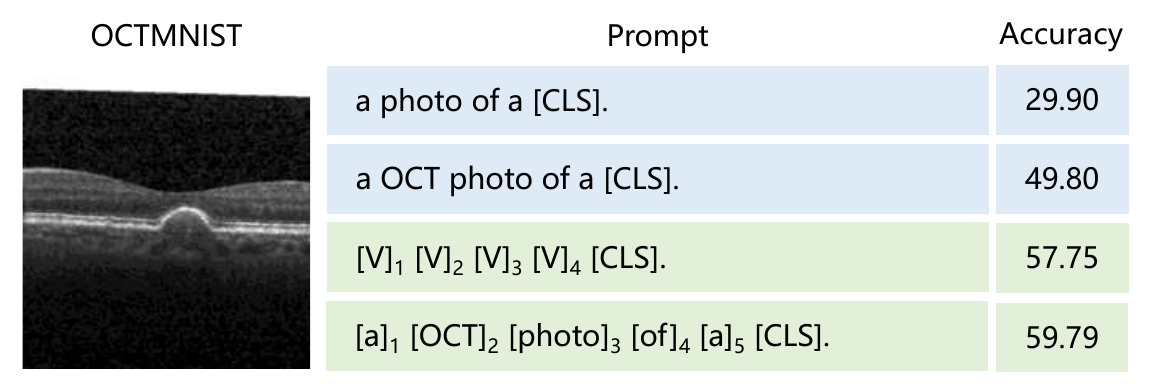}
		\caption{}
		\label{prompt_en_3}%文中引用该图片代号
	\end{subfigure}
        \centering
	\begin{subfigure}{0.49\linewidth}
		\centering
		\includegraphics[width=0.98\linewidth]{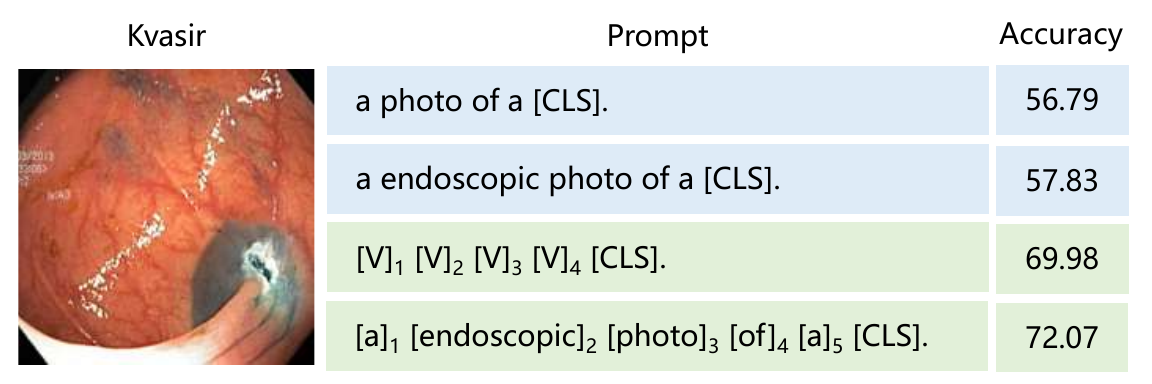}
		\caption{}
		\label{prompt_en_4}%文中引用该图片代号
	\end{subfigure}
        \caption{Comparison of classification accuracies (\%) of different initialization methods. [CLS] denotes [CLASS].}
\end{figure*}

In text understanding, text expressions for natural images typically align well with human experiences, whereas clinical interpretation of pathological features in biomedical images (e.g., space-occupying lesions, exudative changes) must rely on expert biomedical knowledge. This fundamental difference dictates that embedding prior domain-specific knowledge into VLMs is a critical pathway to enhance performance in biomedical image classification. We observe that in downstream classification tasks, the performance of pre-trained VLMs heavily depends on the text prompts. In vocabulary selection, biomedical image analysis demands strict adherence to domain-specific terminology, e.g., simply replacing ``\texttt{photo}'' with ``\texttt{image}'' improved the accuracy of BTMRI~\cite{btmri} (Figure \ref{prompt_en_1}) by 3\%. Effective prompts require domain expertise, e.g., explicitly incorporating modality information in prompts significantly boosts performance, such as ``\texttt{chest X-ray}'' for COVID-QU-Ex~\cite{covid} (Figure \ref{prompt_en_2}); ``\texttt{OCT}'' for OCTMNIST~\cite{octmnist} (Figure \ref{prompt_en_3}); and ``\texttt{endoscopic}'' for Kvasir~\cite{Kvasir} (Figure \ref{prompt_en_4}). Additionally, adjusting the structure of the sentence can also impact the performance, e.g., appending ``\texttt{presented in image}'' after the class label in COVID-QU-Ex (Figure \ref{prompt_en_2}) enhances semantic understanding of the category, increasing the accuracy by 19\%.

Notably, while parameterized text prompts allow for prompt optimization, experiments demonstrate that the initialization of prompts still significantly impacts the accuracy of VLMs. Taking OCTMNIST as an example (Figure \ref{prompt_en_3}), when prompts are adjusted from the generic expression ``\texttt{a photo of a [CLASS].}'' to the domain-specific expression ``\texttt{an OCT photo of a [CLASS].}'', the accuracy can be improved by 3\%. Similar improvements can be observed for other datasets when modality information is incorporated. These findings highlight the substantial impact of domain knowledge on VLMs. Therefore, in the design of biomedical VLMs, using the domain-specific expression and integrating biomedical prior knowledge can effectively ensure the key features localization accurately in biomedical images. This discovery provides important guidance for fine-tuning biomedical VLMs, as embedding domain knowledge can improve their robustness and accuracy in biomedical tasks.

This study addresses the following limitations of current PEFT methods in biomedical imaging applications: 1) in visual representation, existing methods exhibit insufficient sensitivity to subtle pathological features because they suffer from excessive attention in non-diagnostic regions; 2) in text understanding, the lack of expert biomedical knowledge leads to limited accuracy in the clinical semantic interpretation. In order to resolve these critical issues, we propose a knowledge-enhanced dual modality prompt tuning method, with the following main contributions:

\begin{enumerate}
\item In visual localization, we introduce zero vector as an additional soft prompt and employ an layer-wise attention weight adjustment technique to eliminate the unnecessary attention to irrelevant areas so that the method can focus on diagnostically critical regions.

\item In text generation, we construct fixed clinical prompt templates and leverage GPT-4 as a dynamic knowledge engine to generate clinically relevant prompts so that the precision in biomedical semantic comprehension can be improved.

\item In text learning, we utilize Kullback-Leibler (KL)-divergence loss to effectively transfer high-confidence clinical knowledge from GPT-4 to Biomed-DPT, and employ $L_1$ loss as regularization constraints to maintain model generalization capability so that the accuracy of the biomedical imagine classification can be improved.

\end{enumerate}

\section{Related Work}
\subsection{Vision-Language Models}

VLMs, as a crucial paradigm in cross-modal learning, typically consist of three core components: image encoder, text encoder, and loss function. Represented by OpenAI's CLIP~\cite{clip}, which is pre-trained on 400 million internet image-text pairs, the model learns cross-modal semantic alignment by maximizing the similarity of matched image-text pairs while minimizing the similarity of unmatched pairs. The model supports ResNet-50~\cite{resnet} and ViT~\cite{vit} as image encoders and Transformer~\cite{transformer} as the text encoder. These encoders map high-dimensional image and text data into low-dimensional feature vectors, and predictions are generated by computing the cosine similarity between image and text vectors. Due to the pre-trained data covers diverse internet-scale scenarios and concepts, the model learns highly generalizable vision-language association patterns. For example, even if the pre-trained samples never include instances of ``\texttt{retinal lesions}'', the model can still effectively identify such cases through biomedical-related concepts (e.g., ``\texttt{hemorrhage}'' or ``\texttt{exudate}'') learned during pre-training. Notably, although CLIP demonstrates remarkable zero-shot classification capabilities, its performance in biomedical imaging applications remains limited due to the semantic gap between pre-trained data and specialized domains.

In order to overcome the limitations of CLIP in non-natural image domains, several large-scale pre-trained VLMs have been proposed in the biomedical field, including PMC-CLIP~\cite{lin2023pmc}, PubMedCLIP~\cite{pubmedclip}, and BiomedCLIP~\cite{biomedclip}. PMC-CLIP~\cite{lin2023pmc} introduced an automated data construction pipeline and released the PMC-OA dataset, containing 1.6 million high-quality image-caption pairs. However, this study retained the original CLIP architecture without optimizing it for biomedical images. PubMedCLIP~\cite{pubmedclip} performed domain-adaptive fine-tuning based on PubMed literature data, improving only the loss function while making no substantial changes to the model architecture. In contrast, BiomedCLIP~\cite{biomedclip} made advancements in both data scale and model architecture. Its constructed training dataset is two orders of magnitude larger than existing biomedical multimodal datasets. Additionally, to accommodate the characteristics of biomedical data, the model replaced the Transformer with BERT as the text encoder and expanded the context window to handle long texts. It also employed a larger-scale vision encoder (ViT instead of ResNet) to process high-resolution biomedical images, thereby more effectively captures complex semantic relationships in biomedical data. Although these models demonstrate superior performance in foundational task evaluations, they still require fine-tuning for specific downstream tasks to achieve better performance.

\subsection{Few-shot Adaptation of VLMs}
In recent years, VLMs have made significant progress in downstream task adaptation, where prompt learning has gradually replaced traditional fine-tuning methods as the mainstream paradigm in the field of natural image processing due to its parameter-efficient advantages. As pioneering work in this field, CoOp~\cite{coop} and CoCoOp~\cite{cocoop} introduced learnable continuous text prompt vectors, achieving efficient task adaptation while keeping the pre-trained model parameters frozen. However, since the prompt vectors trained on seen-class data in such methods struggle to effectively transfer to out-of-distribution domains, they exhibit notable limitations in domain generalization. In order to address the generalization bottleneck of CoOp, KgCoOp~\cite{kgcoop} mitigated the model's forgetting of general text knowledge by constraining the divergence between learnable prompts and manually designed prompts. Meanwhile, ProGrad~\cite{prograd} employed a gradient alignment strategy to achieve collaborative optimization between task-specific knowledge and pre-trained prior knowledge, both significantly improve the model's cross-domain transfer performance. Notably, while the above methods primarily focus on constraint optimization at the text level, PromptSRC~\cite{promptsrc} proposed a self-regulated prompt method. This approach constructs an expanded text semantic space and regulates prompt representations by maximizing mutual consistency between images and texts with the frozen model, greatly enhanced the model's anti-overfitting capability.

Two key challenges limit conventional prompt learning in biomedicine: first, the sophisticated spatial relationships inherent in medical imaging; second, the critical need for accurate clinical concept representation. These characteristics impose exceptional comprehension demands that necessitate tailored domain adaptation strategies. DCPL~\cite{dcpl} leverages knowledge extracted from the domain foundation model LSDM, using a lightweight network to transform it into domain biases for the vision and language branches. However, DCPL relies on a CLIP-based pre-trained model architecture, which has limited feature representation capability for biomedical images, and its neural network-based fine-tuning requires more computational resources compared to learnable tokens. XCoOp~\cite{xcoop} enhances the adaptability of VLMs in clinical tasks by aligning image semantics, learnable prompts, and clinical concept-driven prompts at multiple granularity levels. While XCoOp employs learnable tokens, it still adopts CLIP as its pre-trained model. BiomedCoOp~\cite{biomedcoop} achieves prompt-context learning by leveraging the semantic consistency of averaged prompt sets from LLMs and a statistics-based prompt selection strategy for knowledge distillation. However, this approach focuses solely on the text level and does not improve fine-tuning methods for biomedical image features. Notably, all these methods rely on LLMs to introduce external domain knowledge. Therefore, effectively mining and utilizing the rich biomedical knowledge embedded in LLMs has become a critical research breakthrough. This direction not only helps bridge the knowledge gap of general models in specialized domains but also provides new technical pathways for developing more precise biomedical image analysis tools.

In the field of few-shot fine-tuning, in addition to prompt learning, adapter learning has also demonstrated significant advantages. CLIP-Adapter~\cite{clipadapter} inserts lightweight fully-connected adapter layers into the image encoder of a pre-trained VLM, requiring only the fine-tuning of a small number of parameters to significantly improve downstream task performance. Tip-Adapter~\cite{tipadapter} proposes a training-free fine-tuning paradigm that eliminates the need to train adapters. Instead, it leverages CLIP's visual encoder to extract visual features from training images, constructing a query-key cache model to obtain adapter weights. Additionally, linear probing~\cite{clip} fine-tunes the model by updating only the parameters of the final linear layer, thus reduced the number of trainable parameters. The enhanced linear probing method, LP++~\cite{lp++}, effectively improves model performance without increasing complexity by fusing multimodal feature representations and introducing a dynamic learning rate adjustment mechanism. These methods collectively form a technical framework for parameter-efficient fine-tuning of VLMs, providing diverse solution options for computationally constrained application scenarios.

\section{Methodology}

\begin{figure*}
    \centering
    \includegraphics[width=0.98\linewidth]{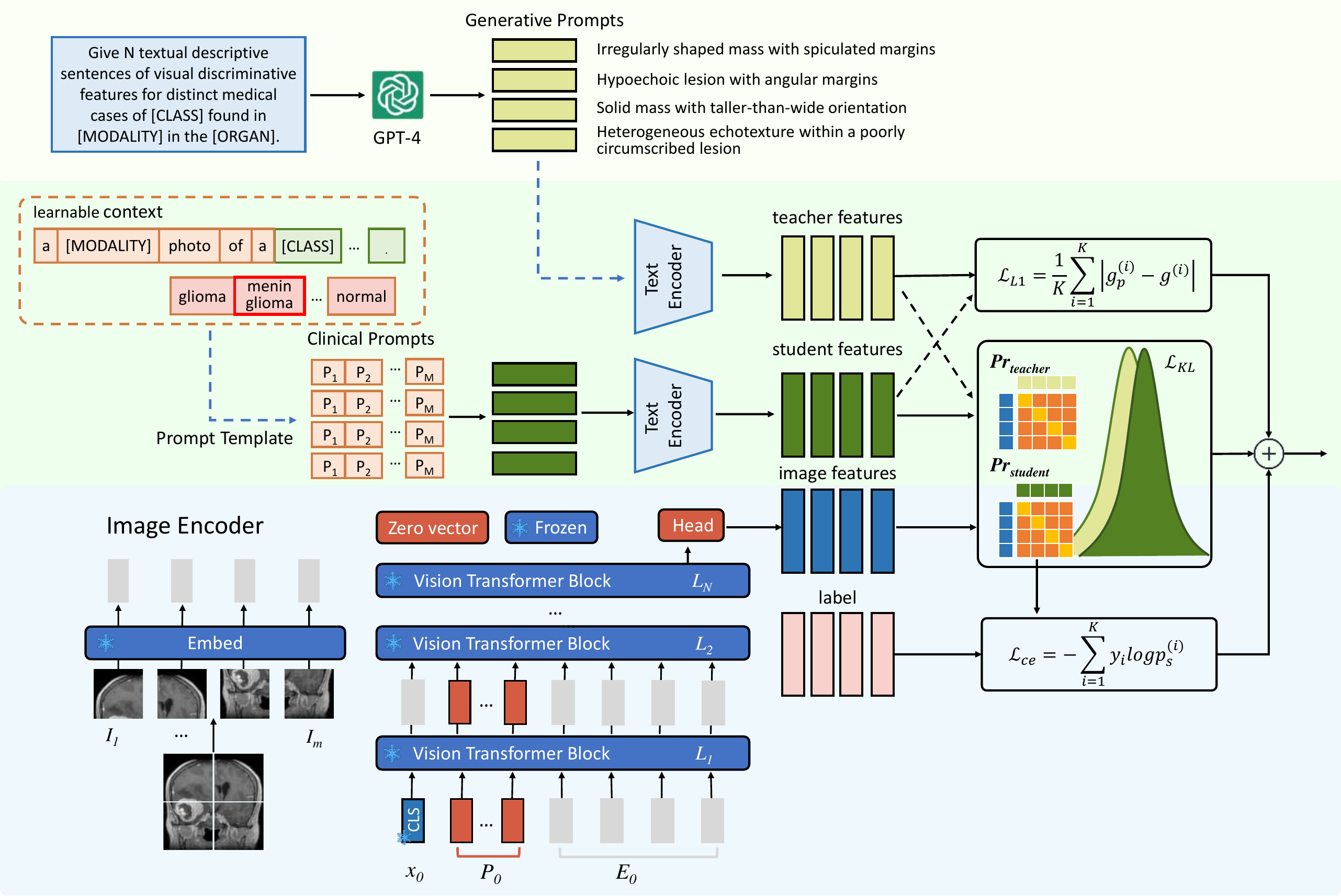}
    \caption{Overview of the Biomed-DPT framework, which combines LLM prompt generation, fixed clinical prompt templates, learnable context, zero vector as a soft prompt, and BiomedCLIP to construct a unified multimodal representation space. In this method, the prompt ensemble strategy is used to integrate text and image features, and the cross-entropy, $L_{1}$ constraint, and KL divergence loss are minimized to achieve effective few-shot learning.}
    \label{fig_FRAME}
\end{figure*}

Figure \ref{fig_FRAME} illustrates the overall framework of the proposed Biomed-DPT, a knowledge-enhanced dual modality prompt tuning method. The approach extracts image and text features based on BiomedCLIP, while introducing multi-layer zero vector as a soft prompts in the image encoder to re-adjust the attention distribution across the image patches, thereby suppressing the excessive attention on non-diagnostic regions. A regularization strategy is designed, employing the $L_1$ norm constraints to enforce the teacher model's supervision over the student model's text features. Additionally, a knowledge distillation component is incorporated, where high-quality clinical diagnostic knowledge generated by GPT-4 is distilled into the student model using the KL divergence. The entire model is optimized through a multi-task objective function combining cross-entropy loss, $L_1$ loss, and KL divergence loss. This ensures the model maintains representation accuracy while achieving enhanced robustness.

\subsection{Clinical knowledge enhancement}
We systematically inject external clinical knowledge into the model by constructing clinical concept-driven prompt templates integrated with diverse prompt words generated by LLMs, thereby optimizing biomedical image classification performance.

\textbf{Fixed Clinical Concept-Driven Prompt Template}. Effective prompt words enhance prompt learning by generating diverse text representations. However, human-designed prompt templates previously used in the natural vision tasks may encounter limitations in the biomedical domain when acquiring relevant clinical knowledge and sufficient descriptive diversity. In order to incorporate biomedical knowledge into the prompt learning process, we design disease-specific prompt templates based on the clinical concepts. These Clinical Prompt Templates are constructed following the principles: each template explicitly states the imaging modality (e.g., MR, X-ray, OCT), specifies the target organ or body part whenever possible, incorporates relevant clinical context for specific diseases (e.g., COVID-QU-Ex), and maintains a consistent template structure by keeping the number of context tokens (nctx) as the word count before \texttt{[CLASS]}.

\begin{table*}[htbp]
\centering
\caption{Clinical prompt template and context parameters in all dataset.}
\label{tab:prompt_templates}
\resizebox{\linewidth}{!}{
\begin{tabular}{llc}
\hline
Datasets & Hand-crafted prompt & nctx \\ 
\hline
BTMRI & \texttt{a MR photo of a [CLASS] in the brain}. & 5 \\
BUSI & \texttt{a ultrasound photo of a [CLASS] in the breast}. & 5 \\
CHMNIST & \texttt{a histopathological photo of a [CLASS]}. & 5 \\
COVID-QU-Ex & \texttt{a chest X-ray photo of a [CLASS] affected by COVID-QU-Ex in the lung}. & 6 \\
CTKIDNEY & \texttt{a CT photo of a [CLASS] in the kidney}. & 5 \\
DermaMNIST & \texttt{a dermatoscopy photo of a [CLASS] in the skin}. & 5 \\
KneeXray & \texttt{a frontal X-ray photo of a [CLASS] in the knee joint}. & 6 \\
Kvasir & \texttt{a endoscopic photo of a [CLASS] in the colon}. & 5 \\
LC25000 & \texttt{a histopathological photo of a [CLASS]}. & 5 \\
OCTMNIST & \texttt{a OCT photo of a [CLASS]}. & 5 \\
RETINA & \texttt{a photo of a [CLASS] presented in image}. & 4 \\
\hline
\end{tabular}
}
\end{table*}

\textbf{Flexible Generative Customized Prompts}. Prompt ensemble significantly enhances the model's generalization capability by integrating diverse descriptions generated by LLMs, capturing lesion characteristics from multiple perspectives (such as morphology, texture, and anatomical location), while reducing cognitive bias from single prompts and improving the model's adaptability to multimodal biomedical images (e.g., X-ray, MRI, ultrasound). Notably, recent studies on GPT-4 have validated its performance in clinical case report-related tasks. We employ LLM GPT-4 to generate image prompts. In order to ensure diversity and relevance in the generated prompts, we use the LLM to produce customized image prompts for each class in the dataset. Specifically, for the C different classes in the dataset, we generate text prompts $X_g\in R^{N\times C\times L}$, where each class is associated with N distinct text outputs ($N=50$). These outputs are generated based on the following query template: ``\texttt{Give $N$ text descriptive sentences of visual discriminative features for distinct biomedical cases of [CLASS] found in [MODALITY] in the [ORGAN].}'' where \texttt{[CLASS]} represents the disease category (e.g., pneumonia, tumor, fracture), \texttt{[MODALITY]} denotes the biomedical imaging modality (e.g., X-ray, CT, MRI), and \texttt{[ORGAN]} specifies the affected organ (e.g., kidney, brain).

\subsection{Learnable Token Text Prompt} 
CLIP learns the alignment relationship between visual and language modalities through pre-training on large-scale image-text paired data, enabling the model to acquire strong zero-shot classification capabilities. This paradigm allows the model to achieve effective classification in downstream tasks without fine-tuning or domain-specific annotations. For a classification task with $K$ categories, CLIP generates text embeddings $W$ through the following process, where we formally define the text encoder $\theta$. CLIP employs a manually designed prompt template ``\texttt{a photo of a [CLASS].}'' as paired text to describe image content, with \texttt{[CLASS]} serving as a placeholder for class labels. The Transformer-based encoder $e(\cdot)$ receives word sequences and outputs vectorized text tokens $t_i = e(\texttt{``a photo of a [CLASS].''})$. However, this fixed handcrafted prompt design in VLMs lacks sufficient generality and struggles to adapt to the specific requirements of different downstream tasks.

In order to address this, we employ continuous learnable context vectors $v = \{v_1, v_2, \dots, v_M\}$ to dynamically generate prompts. $v$ is initialized with $e(\texttt{``a [MODALITY] photo of a [CLASS].''})$, where $M$ denotes the number of corresponding prompt tokens and \texttt{[MODALITY]} represents the biomedical imaging modality (e.g., X-ray, CT, MRI). The class name encoding $c_i$ for the $i$-th class is embedded into the context vector $v$ to generate the prompt $t_i = \{v_1, v_2, \dots, v_M, c_i\}$. The text encoder $\theta(\cdot)$ takes the vectorized text tokens $t_i$ of the $i$-th class as input and outputs the text class embedding $\theta(\cdot)$. The set of text embeddings for all classes is denoted as $W = \{w_i\}_{i=1}^{K}$.

\subsection{Zero-vector As A Soft Visual Prompt} 
While text prompt tuning in VLMs enhances cross-task transferability by effectively focusing on key regions, it may lead to excessive reliance on the class token. Therefore, we insert a Zero-vector after the class token to serve as a Soft Prompt, enabling the class token to aggregate information and distribute the model's attention. This approach allows the model to utilize the balanced information from all image patches.

For an input image $\mathbf{I} \in \mathbb{R}^{3\times H\times W}$, we first partition it into $N_p$ local patches $\{\mathbf{I}_j \in \mathbb{R}^{3\times h\times w} \mid j=1,...,N_p\}, \quad N_p = \left\lfloor \frac{H}{h} \right\rfloor \times \left\lfloor \frac{W}{w} \right\rfloor$, where $h \times w$ represents the resolution of a single patch. The embedding layer then linearly projects each image patch $\mathbf{I}_j$ into a $d$-dimensional embedding space:
\begin{equation}
   e_0^j = \text{Embed}(I_j), \quad e_0^j \in \mathbb{R}^d, \quad j = 1, 2, \dots, N_p.
\end{equation}

Let $E_l = \{ e_l^j\in \mathbb{R}^d \mid j \in \mathbb{N}, 1 \leq j \leq N_p \}$ and $P_l = \{ 0_l^i \in \mathbb{R}^d \mid i \in \mathbb{N}, 1 \leq i \leq P \}$ denote the image embeddings and visual prompts at the $l$-th Transformer layer respectively, where $x_l$ represents a class token in the image encoder. We introduce zero vector as a soft prompt in each transformer block $\Phi_l$:
\begin{equation}
[x_l, P_l, E_l] = \Phi_l ([x_{l-1}, P_{l-1}, E_{l-1}]), \quad l = 1, 2, \cdots, L.
\end{equation}

Given an image $I$ and its label $y$, we employ a visual encoder $\phi(\cdot)$ to extract visual embeddings. The entire image feature extraction process can be simply expressed as:
\begin{equation}
\mathbf{f} = \phi([x_0, P_0, \dots , P_L, E_0])
\end{equation}

The model's prediction probability is calculated by the following:

\begin{equation}\label{eq:logits}
p(y = i \mid \mathbf{x}) = \frac{\exp\left(\mathrm{sim}(w_y, \mathbf{f}) / \tau\right)}{\sum_{i=1}^K \exp\left(\mathrm{sim}(w_i, \mathbf{f}) / \tau\right)}
\end{equation}

\begin{equation}
\mathrm{sim}(w_i, \mathbf{f}) = \frac{w_i^\top \mathbf{f}}{\|w_i\| \cdot \|\mathbf{f}\|}
\end{equation}

where $\mathbf{f} \in \mathbb{R}^d$ represents the feature vector extracted by the image encoder, $w_i \in \mathbb{R}^d$ denotes the weight vector generated by the text encoder for the extended text of the $i$-th class, $\mathrm{sim}(\cdot)$ indicates the similarity function, $K$ is the total number of classes, and $\tau$ is the temperature parameter learned by CLIP to adjust the smoothness of the probability distribution. The final prediction result of the model is given by:
\begin{equation}\label{eq:pred}
\hat{y} = \arg\max_i p(y = i \mid \mathbf{x})
\end{equation}

In the computation process of the text encoder, the dynamically generated learnable context vectors can automatically extract semantic features highly relevant to the current task, avoiding the prior bias introduced by manual prompt design. In the computation process of the image encoder, zero vectors dynamically adjust the attention weight distribution among image patches in the self-attention layer, reducing excessive reliance on local features. In the context of biomedical image analysis, this Dual Modality Prompt Tuning enables the model to simultaneously capture fine-grained visual features of lesions and clinically relevant semantic contexts, thereby achieving precise alignment of vision-language representations.

\textbf{Learning the Multi-objective Function.} In order to mitigate the semantic discrepancy between prompt features and LLM-generated domain-specific features, we employ a multi-objective function including the $L_1$ regularization loss, KL divergence loss, and cross-entropy loss. We implement the constrained optimization at the feature level through a feature distance minimization strategy to reduce distribution differences between general features and prompt features, thereby enhancing the robustness of feature representation. At the distribution level, the KL divergence loss is utilized for probability distribution matching to facilitate efficient cross-model knowledge transfer. At the task level, cross-entropy loss optimizes classification boundaries to ensure fundamental classification performance. This multi-objective optimization framework achieves comprehensive alignment from feature space to prediction space through a hierarchical constraint mechanism. 

At the feature level, to ensure semantic consistency between text-encoded features and domain-customized prompts, we establish an $L_1$-norm based feature distance constraint. Given the original features $g\in \mathbb{R}^d$ extracted by the text encoder and the customized prompt features $g_p\in \mathbb{R}^d$ generated by GPT-4, the $L_1$ loss is defined as:
\begin{equation}
\mathcal{L}_{L_1}=\frac{1}{K}\sum_{i=1}^K|g_p^i-g^i|
\end{equation}
where $K$ denotes the number of classes. By utilizing the $L_1$ loss, the text-encoded features can converge toward the customized prompt features in the embedding space, thereby enhancing the model's ability to comprehend domain-specific semantics.

At the distribution level, to further enhance prediction-level consistency, we define the KL divergence loss:
\begin{equation}
\mathcal{L}_{KL}(\mathbf{p}_t \parallel \mathbf{p}_s) = \sum_{i=1}^K p_t^{(i)} \log \frac{p_t^{(i)}}{p_s^{(i)}}
\end{equation}

where the teacher distribution $\mathbf{p}_t$ and student distribution $\mathbf{p}_s$ are calculated by Equation \ref{eq:logits}. By optimizing the KL divergence loss function, we maintain consistency between the GPT-prompted logits distribution and the pre-trained BiomedCLIP logits distribution.

At the task level, to ensure model classification performance, we define the cross-entropy loss between predictions $\mathbf{p}_s$ and ground truth labels $\mathbf{y}$ as:
\begin{equation}
\mathcal{L}_{ce}(\mathbf{y}, \mathbf{p}_s) = -\sum_{i=1}^K y_i \log p_s^{(i)}\label{eq:ce}
\end{equation}

During training, we jointly optimize the regularization constraint $\mathcal{L}_{L_1}$, KL divergence loss $\mathcal{L}_{KL}$, and cross-entropy loss $\mathcal{L}_{ce}$ to maximize mutual consistency between prompt features and LLM-generated domain-specific features. The total loss function is defined as:

\begin{equation}
\mathcal{L}=\mathcal{L}_{ce}+\lambda_1 \mathcal{L}_{L_1}+\lambda_2 \mathcal{L}_{KL}
\end{equation}
where $\lambda_1$ and $\lambda_2$ are weighting coefficients used to balance the contributions of each loss term.

\section{Experiments and Results}\label{sec_result}
\subsection{Experimental Setup}
In order to evaluate the performance of Biomed-DPT, we first compare different initialization templates for CoOp and Biomed-DPT, then conduct assessments from the following three aspects: 1) Few-Shot Evaluation, evaluating the classification accuracy of the model with limited annotated data; 2) Base-to-Novel Class Generalization, examining the model's transfer performance on unseen classes; 3) Ablation Experiments, including analysis of the contributions of each component, comparison of different number of LLM prompts, comparison of different pre-trained VLMs, and visualization of how the model localizes lesion regions. In order to ensure the reliability and comparability of the experimental results, all experiments are conducted using pre-trained BiomedCLIP.

\textbf{Implementation details:}
We conduct the experiments using ViT-B/16~\cite{vit}-based visual encoder and BERT~\cite{bert}-based text encoder. Prompt templates as shown in Table~\ref{tab:prompt_templates} are used to initialize the context vectors. In order to ensure the reliability and fairness of the experimental results, the final accuracy is obtained by averaging the results across three different random seeds such as seed 1, 2, 3. For the training setup, we employe 100 epochs for few-shot learning experiments and 50 epochs for the Base-to-Novel classes benchmark. During dataset processing, we utilize 50 prompts generated by LLM GPT-4. The learning rate is set to 0.0025, with batch size of 4, and the model is trained using the SGD optimizer. The hyperparameters $\lambda_1$, $\lambda_2$ are both set to their optimal values. All experiments are conducted on three RTX 4090 GPUs (24 GB RAM each).

\textbf{Comparative methods:}
This study selects four representative types of pre-trained VLMs for comparison, including CLIP~\cite{clip} pre-trained on natural images, PMC-CLIP~\cite{lin2023pmc} pre-trained on the PMC-OA dataset, PubMedCLIP~\cite{pubmedclip} fine-tuned via domain adaptation on PubMed literature, and BiomedCLIP~\cite{biomedclip} pre-trained on the PMC-15M dataset. Specifically, CLIP~\cite{clip} achieves cross-modal semantic alignment through contrastive learning, but its performance in the biomedical domain is limited due to the distributional gap between natural and biomedical images. PMC-CLIP~\cite{lin2023pmc} is pre-trained on the PMC-OA dataset, but it only replaces the training data without modifying the architecture. Meanwhile, PubMedCLIP~\cite{pubmedclip} employs weighted cross-entropy loss for domain-adaptive fine-tuning on PubMed literature, yet it remains constrained by the original CLIP architecture, hindering its ability to fully capture the complex semantics of biomedical texts. In order to overcome this limitation, BiomedCLIP~\cite{biomedclip} adopts BERT-based text encoder with an extended context window while enhancing the visual encoder to handle high-resolution biomedical images, achieving superior cross-modal alignment.

At the methodological level, this study compares multiple cutting-edge prompt learning and fine-tuning strategies, including two linear probing methods (Standard LP~\cite{clip}, LP++~\cite{lp++}), two CLIP-based adapter approaches (CLIP-Adapter~\cite{clipadapter}, Tip-Adapter~\cite{tipadapter}), and five text prompt learning methods (CoOp~\cite{coop}, CoCoOp~\cite{cocoop}, KgCoOp~\cite{kgcoop}, ProGrad~\cite{prograd}, BiomedCoOp~\cite{biomedcoop}). Specifically, Linear Probing~\cite{clip} fine-tunes only the last linear layer, while LP++~\cite{lp++} introduces multimodal feature fusion and dynamic learning rates on top of this, though the expressive power of the linear layer remains limited. CLIP-Adapter~\cite{clipadapter} achieves PEFT by inserting lightweight adapters, whereas Tip-Adapter~\cite{tipadapter} further proposes a training-free cache model approach, yet neither fully leverages text modality information.CoOp~\cite{coop} pioneers learnable text prompts but suffers from static prompts that struggle to adapt to different samples. CoCoOp~\cite{cocoop} addresses this limitation with instance-specific prompts but risks deviating from pre-trained knowledge. In order to mitigate this, KgCoOp~\cite{kgcoop} constrains prompt divergence from manually designed templates to preserve general knowledge, while ProGrad~\cite{prograd} innovatively employs gradient alignment to enable cooperative optimization of old and new knowledge. BiomedCoOp~\cite{biomedcoop} integrates domain expertise from LLMs and constrains model generalization via Euclidean distance. Additionally, we evaluate BiomedCLIP's zero-shot capability and its prompt ensemble variant.

\textbf{Datasets:}
As shown in Table~\ref{tab:datasets}, we conduct experiments on 11 different biomedical imaging datasets, which cover 10 distinct organs and 9 imaging modalities, including computed tomography (CTKidney~\cite{ctkidney}, for kidney imaging), dermoscopy (DermaNIST~\cite{dermamnist, dermamnist2}, for skin lesion analysis), endoscopy (Kvasir~\cite{Kvasir}, for gastrointestinal disease detection), fundus photography (RETINA~\cite{retina2}, for retinal disease diagnosis), histopathology (LC2500~\cite{lc25000} and CHMNIST~\cite{chmnist}, for tissue slice analysis), magnetic resonance imaging (BTMRI~\cite{btmri}, for brain tumor detection), optical coherence tomography (OCTMNIST~\cite{octmnist}, for retinal tomography), ultrasound (BUSI~\cite{busi}, for breast lesion detection), and X-Ray (COVID QU-Ex~\cite{covid} and KneeXray~\cite{kneexray}, for COVID-QU-Ex diagnosis and knee lesion analysis, respectively).

\begin{table}[!t]
\centering
\caption{An overview of the 11 datasets used spanning 9 biomedical imaging modalities and 10 different organs.}
\label{tab:datasets}
\begin{tabular}{lllc}
\hline
Datasets & Modality & Organ & Classes \\ 
\hline
BTMRI~\cite{btmri} & MRI & Brain & 4 \\
BUSI~\cite{busi} & Ultrasound & Breast & 3 \\
CHMNIST~\cite{chmnist} & Histopathology & Colorectum & 8 \\
COVID-QU-Ex~\cite{covid} & X-ray & Chest & 4 \\
CTKIDNEY~\cite{ctkidney} & CT & Kidney & 4 \\
DermaMNIST~\cite{dermamnist, dermamnist2} & Dermatoscopy & Skin & 7 \\
KneeXray~\cite{kneexray} & X-ray & Knee & 5 \\
Kvasir~\cite{Kvasir} & Endoscopy & Colon & 4 \\
LC25000~\cite{lc25000} & Histopathology & Lung Colon & 6 \\
OCTMNIST~\cite{octmnist} & OCT & Retina & 4 \\
RETINA~\cite{retina2} & Fundus Photography & Retina & 4 \\
\hline
\end{tabular}
\end{table}

\subsection{Prompt Template Evaluation}
In this section, we investigate the impact of different initialization templates on the performance of fine-tuning methods, including random initialization ($[V]_{1}[V]_{2}[V]_{3}[V]_{4}$), natural prompt template initialization (``\texttt{a photo of a [CLASS]}.''), and clinical prompt template initialization (CPT) (see Table~\ref{tab:prompt_templates}). We conducted comparative experiments using five different training scales ($K=1, 2, 4, 8, 16$) for both context optimization (CoOp) and Biomed-DPT. The results are presented in Table~\ref{tab:init_methods}.

Different prompt initialization strategies exhibit significant performance disparities in few-shot learning scenarios. Among these, CPT initialization consistently achieves the highest accuracy across all sample sizes. As the number of training samples $K$ increases from 1 to 16, all initialization methods show progressive performance gains. And when K=16, CPT initialization maintains consistent advantages of 0.96\% for CoOp and 0.95\% for Biomed-DPT, demonstrating its scalability with larger datasets. Across all sample sizes, this approach yields an average accuracy improvement of 0.73\% over the suboptimal natural prompt template in CoOp, and a more pronounced 1.06\% enhancement in Biomed-DPT. 

\begin{table}[htbp]
\centering
\caption{Classification accuracy under different initialization prompt methods(\%).}
\label{tab:init_methods} 
\begin{tabular}{lccccc}
\toprule
Method & $K=1$ & $K=2$ & $K=4$ & $K=8$ & $K=16$ \\ 
\midrule

\rowcolor{gray}
\multicolumn{6}{c}{\textbf{CoOp}} \\
$[V]_{1}[V]_{2}[V]_{3}[V]_{4}$ & 50.74 & 54.78 & 58.22 & 63.28 & 65.93 \\
``a photo of a'' & 50.18 & 54.17 & 59.77 & 65.85 & 69.72 \\
CPT & 50.90 & 54.68 & 61.28 & 65.79 & 70.68 \\
\midrule

\rowcolor{gray}
\multicolumn{6}{c}{\textbf{Biomed-DPT}} \\
$[V]_{1}[V]_{2}[V]_{3}[V]_{4}$ & 55.25 & 59.31 & 61.71 & 65.88 & 70.12 \\
``a photo of a'' & 56.99 & 60.68 & 65.37 & 69.81 & 72.56 \\
CPT & \textbf{59.03} & \textbf{61.27} & \textbf{66.12} & \textbf{70.76} & \textbf{73.51} \\
\bottomrule
\end{tabular}
\vspace{0.2cm}
\begin{flushleft}
\footnotesize 
\end{flushleft}
\end{table}

These experimental results demonstrate that the domain-knowledge-integrated clinical prompt template can effectively guide VLMs to capture key features in biomedical images. The findings not only verify the significant impact of prompt initialization on model performance, but more importantly highlight the value of domain-adapted prompts in biomedical image analysis.

\subsection{Few-shot Evaluation}
In this section, we compare Biomed-DPT with various advanced prompt learning methods, including five text prompt learning methods (CoOp, CoCoOp, KgCoOp, ProGrad, BiomedCoOp), two CLIP-based adapter methods (CLIP-Adapter, Tip-Adapter), and two linear probing methods (Standard LP, LP++). We use zero-shot BiomedCLIP and LLM-prompted BiomedCLIP as benchmarks. All comparative methods base on BiomedCLIP and are trained under optimal parameter settings.

As shown in Table \ref{tab:avg_method}, Biomed-DPT demonstrates significant advantages across all experimental settings, particularly excelling in low-shot scenarios. Specifically, under the conditions of $K=1, 2, 4, 8, 16$, Biomed-DPT outperforms the suboptimal method BiomedCoOp by 2.34\%, 2.69\%, 1.69\%, 1.92\%, and 1.38\%, respectively. Experimental observations reveal that CLIP-Adapter exhibits a accuracy decline trend after 1-shot, while methods such as Standard LP, KgCoOp show diminishing accuracy improvements as the sample size increases. In order to address this phenomenon, we mitigate overfitting in few-shot settings by introducing an $L_1$ regularization strategy. Notably, Biomed-DPT achieves an accuracy improvement of approximately 17\% over zero-shot BiomedCLIP with just 1-shot training. When the training samples are expanded to 16-shot, the accuracy advantage $\Delta_{\text{acc}}$ further increases to about 24\%, fully demonstrating the method's exceptional performance in biomedical image classification tasks.

\begin{table}[htbp]
\centering
\caption{Evaluation against state-of-the-art techniques. This table presents the average classification accuracy (\%) obtained from 11 benchmarks, along with mean from 3 sampled support sets for each dataset. The top-performing results are in bold.}
\label{tab:avg_method}
\begin{tabular}{lccccc}
\toprule
Method & $K=1$ & $K=2$ & $K=4$ & $K=8$ & $K=16$ \\ 
\midrule
\rowcolor{gray}
\multicolumn{6}{c}{\textbf{Zero-shot Methods}} \\
BiomedCLIP~\cite{biomedclip} & \multicolumn{5}{c}{42.05}  \\
\makecell[l]{BiomedCLIP~\cite{biomedclip}\\+Ensemble}  & \multicolumn{5}{c}{56.02}   \\
\midrule
\rowcolor{gray}
\multicolumn{6}{c}{\textbf{CLIP-based Adapter Methods}} \\
CLIP-Adapter~\cite{clipadapter} & 46.54 & 45.18 & 45.45 & 46.27 & 47.75 \\
Tip-Adapter~\cite{tipadapter} & 49.11 & 54.06 & 58.93 & 63.46 & 67.18 \\
\midrule
\rowcolor{gray}
\multicolumn{6}{c}{\textbf{Linear Probing Methods}} \\
Standard LP~\cite{clip} & 51.80 & 54.54 & 60.49 & 67.56 & 68.76 \\
LP++~\cite{lp++} & 53.57 & 53.55 & 57.26 & 64.89 & 68.70 \\
\midrule
\rowcolor{gray}
\multicolumn{6}{c}{\textbf{Prompt Learning Methods}} \\
CoOp~\cite{coop} & 50.18 & 54.17 & 59.77 & 65.85 & 69.72 \\
CoCoOp~\cite{cocoop} & 48.53 & 51.28 & 54.69 & 61.09 & 65.10 \\
KgCoOp~\cite{kgcoop} & 52.16 & 54.24 & 59.21 & 63.87 & 64.85 \\
ProGrad~\cite{prograd} & 51.37 & 54.40 & 60.61 & 65.50 & 67.03 \\
BiomedCoOp~\cite{biomedcoop} & 56.69 & 58.58 & 64.45 & 68.84 & 72.13 \\
\rowcolor{purple}
\makecell[l]{Biomed-DPT(Ours)} & \textbf{59.03} & \textbf{61.27} & \textbf{66.12} & \textbf{70.76} & \textbf{73.51} \\
\bottomrule
\end{tabular}
\end{table}

\begin{figure}
    \centering
    \includegraphics[width=0.9\linewidth]{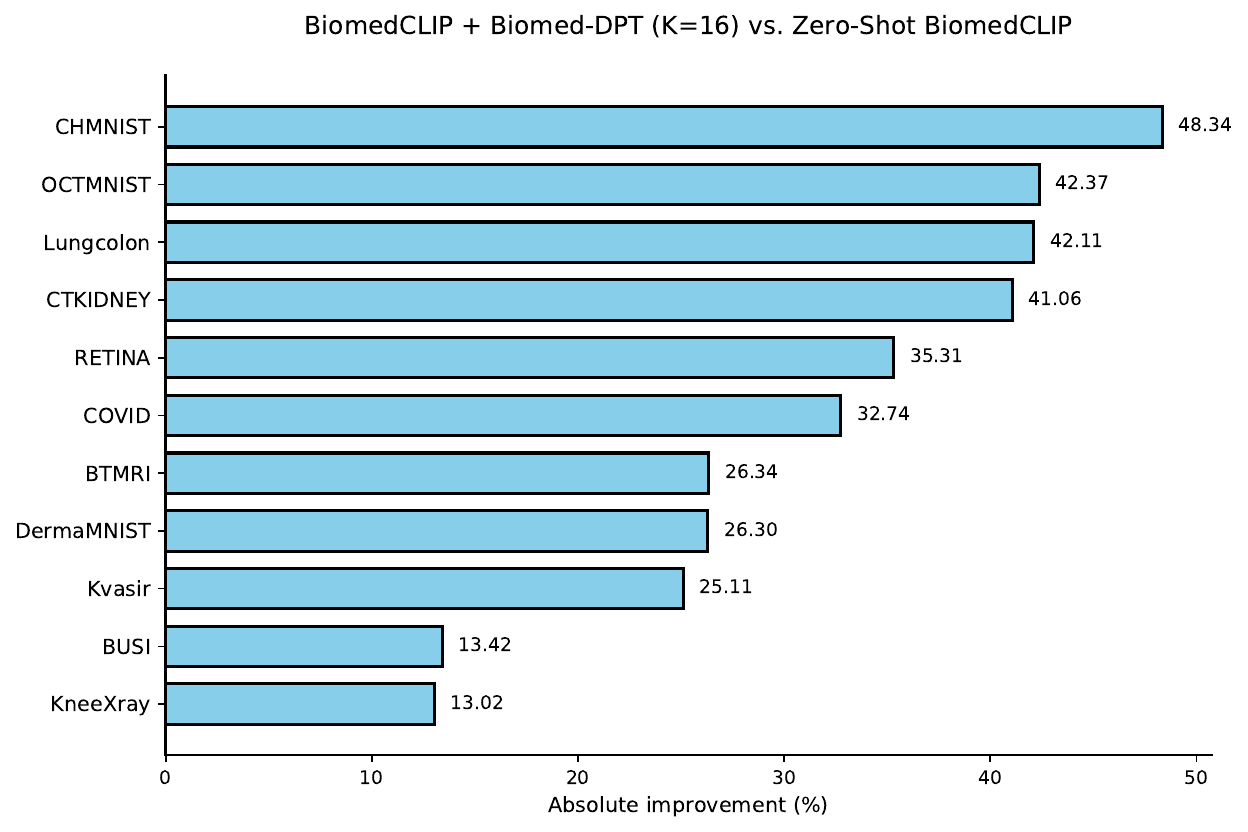}
    \caption{Comparison with hand-crafted prompts ``\texttt{a photo of a [CLASS]}''.}
    \label{absolute}
\end{figure}

Figure \ref{absolute} presents the ranking of absolute accuracy improvements achieved by Biomed-DPT in the 16-shot setting compared to the zero-shot BiomedCLIP with natural prompt template ``\texttt{a photo of a [CLASS]}''. Our default model is BiomedCLIP+Biomed-DPT with $K=16$ and the class token positioned at the end. Notably, Biomed-DPT achieves remarkable accuracy improvements exceeding 40\% on CHMNIST, LC25000, OCTMNIST, and CTKIDNEY. Furthermore, the accuracy gains improved over 20\% are observed on most datasets, including RETINA, COVID, BTMRI, et al. Even on the lowest-performing KneeXray dataset, an improvement of over 13\% is achieved.

\subsection{Base-to-Novel Generalization}
In this section, we divide each dataset into two groups including base classes and novel classes. In order to evaluate the generalization capability of CoOp-based methods, we compare our approach with five advanced text prompt learning methods (CoOp, CoCoOp, KgCoOp, ProGrad, and BiomedCoOp) by conducting prompt tuning on base classes and then evaluating on novel classes. The detailed results are presented in Table \ref{basetonew}, which summarizes the average performance across all 11 datasets under 16-shot settings. It is worth noting that due to insufficient diversity in the novel classes samples of the BUSI dataset, all methods achieve 100\% accuracy on its novel classes.

As shown in Table \ref{basetonew}, Biomed-DPT achieves superior average accuracy across the 11 benchmark datasets compared to existing methods. Specifically, it outperforms the suboptimal method BiomedCoOp by 1.90\% in base class and by 2.88\% in novel classes. For base classes, our method achieves the best performance on 9/11 datasets, while BiomedCoOp only ranks first on 1/11 datasets. For novel classes, our method attains the highest accuracy on 7/11 datasets, whereas BiomedCoOp leads on only 3/11 datasets. The performance advantage is particularly pronounced on specific datasets, the novel classes accuracy on CHMNIST and KneeXray surpasses BiomedCoOp by 9.27\% and 9.85\%, respectively. The superior performance on novel classes demonstrates that Biomed-DPT possesses exceptional cross-category generalization capability, enabling effective adaptation to unseen biomedical categories. This characteristic is particularly crucial in biomedical scenarios where disease patterns continuously evolve.

\begin{table*}[htbp]
\centering
\caption{Accuracy comparison (\%) on Base-to-Novel generalization of Biomed-DPT with SOTA prompt learning methods. HM = harmonic mean of the classification accuracy between base and novel classes.}
\label{basetonew}
\begin{tabular}{ll|cccccc}
\toprule
Dataset & & \makecell[c]{CoOp\\~\cite{coop}} & \makecell[c]{CoCoOp\\~\cite{cocoop}} & \makecell[c]{KgCoOp\\~\cite{kgcoop}} & \makecell[c]{ProGrad\\~\cite{prograd}} & \makecell[c]{BiomedCoOp\\~\cite{biomedcoop}} & \makecell[c]{Biomed-DPT\\(Ours)} \\
\midrule
\multirow{3}{*}{\makecell[l]{Average on \\11 datasets}}  & Base & 74.28 & 72.95 & 69.59 & 71.97 & 76.16 & \textbf{78.06} \\
& Novel & 67.93 & 70.02 & 67.15 & 69.06 & 73.09 & \textbf{75.97} \\
& HM & 71.11 & 71.48 & 68.37 & 70.52 & 74.62 & \textbf{77.02} \\
\midrule
\multirow{3}{*}{BTMRI} & Base & 82.25 & 77.88 & 78.16 & 81.25 & 82.21 & \textbf{83.21} \\
& Novel & 94.51 & 94.84 & 94.94 & 94.29 & \textbf{96.44} & 95.42 \\
& HM & 88.38 & 86.36 & 86.55 & 87.77 & \textbf{89.33} & 89.32 \\
\midrule
\multirow{3}{*}{BUSI} & Base & 78.63 & 79.83 & 79.31 & 78.12 & 76.92 & \textbf{81.54} \\
& Novel & 100.00 & 100.00 & 100.00 & 100.00 & 100.00 & 100.00 \\
& HM & 89.32 & 89.92 & 89.66 & 89.06 & 88.46 & \textbf{90.77} \\
\midrule
\multirow{3}{*}{CHMNIST} & Base & 89.41 & 87.77 & 75.84 & 82.89 & 89.09 & \textbf{89.89} \\
& Novel & 35.11 & 42.51 & 38.88 & 44.81 & 45.21 & \textbf{54.48} \\
& HM & 62.26 & 65.14 & 57.36 & 63.85 & 67.15 & \textbf{72.18} \\
\midrule
\multirow{3}{*}{COVID-QU-Ex} & Base & 75.92 & 77.27 & 75.66 & 75.53 & 74.88 & \textbf{76.34} \\
& Novel & 90.06 & 87.63 & 89.56 & 90.57 & 90.38 & \textbf{91.10} \\
& HM & 82.99 & 82.45 & 82.61 & 83.05 & 82.63 & \textbf{83.72} \\
\midrule
\multirow{3}{*}{CTKIDNEY} & Base & 82.26 & 81.96 & 80.98 & 82.51 & \textbf{86.23} & 82.22 \\
& Novel & 67.92 & 56.56 & 59.89 & 57.48 & \textbf{77.15} & 77.04 \\
& HM & 75.09 & 69.26 & 70.44 & 70.00 & \textbf{81.69} & 79.63 \\
\midrule
\multirow{3}{*}{DermaMNIST} & Base & 48.06 & 42.88 & 36.97 & 35.36 & 58.49 & \textbf{62.30} \\
& Novel & 59.41 & 60.64 & 46.18 & 59.34 & \textbf{55.12} & 48.23 \\
& HM & 53.74 & 51.76 & 41.58 & 47.35 & \textbf{56.80} & 55.26 \\
\midrule
\multirow{3}{*}{KneeXray} & Base & 38.25 & 34.08 & 37.00 & 41.17 & 42.65 & \textbf{51.32} \\
& Novel & 47.69 & 63.02 & 58.88 & 57.06 & 72.63 & \textbf{82.48} \\
& HM & 42.97 & 48.55 & 47.94 & 49.12 & 57.64 & \textbf{66.90} \\
\midrule
\multirow{3}{*}{Kvasir} & Base & 86.22 & 85.94 & 82.22 & 82.72 & 86.83 & \textbf{87.06} \\
& Novel & 58.06 & 53.95 & 58.67 & \textbf{60.22} & 59.11 & 59.22 \\
& HM & 72.14 & 69.94 & 70.44 & 71.47 & 72.97 & \textbf{73.14} \\
\midrule
\multirow{3}{*}{LC25000} & Base & 90.12 & 88.33 & 89.52 & 90.97 & 92.90 & \textbf{94.51} \\
& Novel & 87.57 & 95.02 & 86.56 & 87.59 & 94.10 & \textbf{96.44} \\
& HM & 88.84 & 91.68 & 88.04 & 89.28 & 93.50 & \textbf{95.48} \\
\midrule
\multirow{3}{*}{OCTMNIST} & Base & 75.00 & 79.60 & 68.67 & 72.73 & \textbf{78.93} & 77.00 \\
& Novel & 50.00 & 50.47 & 50.00 & 50.00 & 50.40 & \textbf{63.00} \\
& HM & 62.50 & 65.04 & 59.34 & 61.36 & 64.67 & \textbf{70.00} \\
\midrule
\multirow{3}{*}{RETINA} & Base & 70.98 & 66.88 & 61.14 & 68.41 & 68.62 & \textbf{73.25} \\
& Novel & 56.90 & 65.56 & 55.07 & 58.32 & 63.41 & \textbf{68.29} \\
& HM & 63.94 & 66.22 & 58.10 & 63.36 & 66.02 & \textbf{70.77} \\
\bottomrule
\end{tabular}
\end{table*}

\subsection{Ablation experiments}
\subsubsection{Effect of Different Components}

This section evaluates the contribution of different components in Biomed-DPT through ablation studies, including zero vector as a Soft Prompts (ZSP), Clinical Prompt Template (CPT), $L_{1}$ Loss ($\mathcal{L}_{L_{1}}$), and KL-divergence Loss ($\mathcal{L}_{KL}$). The experimental results are shown in Table~\ref{ablation}.

The baseline CoOp achieves 74.28\% accuracy on base classes but only 67.93\% on novel classes, indicating its limited generalization capability for unseen categories. By incorporating the ZSP, the model shows a 3.45\% improvement in novel classes and a 0.65\% boost in few-shot average accuracy, verifying that dispersing background attention effectively enhances model generalization. Adding the CPT increases few-shot average accuracy by 0.98\%, but causes a 0.98\% drop in novel classes accuracy, suggesting that CPT benefits few-shot learning while potentially introducing overfitting risks. Subsequent integration of the $\mathcal{L}_{L_{1}}$ regularization strategy recovers novel classes accuracy to 71.28\%. The model achieves optimal performance after incorporating the knowledge distillation loss function $\mathcal{L}_{KL}$.

In summary, Biomed-DPT achieves comprehensive performance improvements through, ZSP can prevent excessive focus on non-diagnostic features, CPT can enhance key feature extraction, $\mathcal{L}_{1}$ regularization can suppress model overfitting, and KL-divergence loss $\mathcal{L}_{KL}$ for domain knowledge transfer to boost few-shot learning capability. The final model demonstrates a 3.78\% improvement in base classes accuracy while achieving 8.04\% higher novel classes accuracy and 6.20\% better few-shot average classification accuracy.

\begin{table*}[htbp]
\caption{Impact of each component of Biomed-DPT on the accuracy (\%) of few-shot and Base-to-Novel benchmarks, including ZSP, CPT, $\mathcal{L}_{L_{1}}$, $\mathcal{L}_{KL}$.}
\label{ablation}
\centering
\begin{tabular}{lccccccccc}
\toprule
\multirow{2}{*}{Components} & \multicolumn{3}{c}{Base-to-Novel} & \multicolumn{6}{c}{Few-shot} \\
\cmidrule(lr){2-4} \cmidrule(lr){5-10}
 & Base & Novel & HM & 1 & 2 & 4 & 8 & 16 & Avg \\
\midrule
CoOp~\cite{coop} & 74.28 & 67.93 & 71.11 & 50.18 & 54.17 & 59.77 & 65.85 & 69.72 & 59.94 \\
+ZSP & 76.40 & 71.38 & 73.89 & 50.61 & 54.51 & 60.84 & 66.99 & 70.01 & 60.59  \\
+ZSP+CPT & 76.64 & 70.40 & 73.52 & 53.13 & 55.35 & 61.73 & 67.56 & 70.09 & 61.57 \\
+ZSP+CPT+$\mathcal{L}_{L_{1}}$ & 75.83 & 71.28 & 73.56 & 53.56 & 56.36 & 62.39 & 68.22 & 70.38 & 62.18 \\
Biomed-DPT(+ZSP+CPT+$\mathcal{L}_{L_{1}}$+$\mathcal{L}_{KL}$)  & \textbf{78.06} & \textbf{75.97} & \textbf{77.02} & \textbf{59.03} & \textbf{61.27} & \textbf{66.12} & \textbf{70.76} & \textbf{73.51} & \textbf{66.14} \\
\bottomrule
\end{tabular}
\end{table*}
\subsubsection{Effect of Number of LLM Prompts}
This section investigates the impact of varying numbers of LLM-generated prompts on the performance of the Biomed-DPT model under different few-shot learning scenarios. As shown in Table~\ref{dif_prompt}, results demonstrate an overall performance improvement as prompt quantity increases from 10 to 50. The most significant improvement occurs in the zero-shot condition ($K=0, +5.04\%$), whereas the effect diminishes considerably in the 16-shot setting ($K=16, +3.45\%$), indicating that increasing prompt quantity yields greater marginal benefits in low-sample scenarios. Notably, within the specific transition from 30 to 40 prompts, model accuracy exhibits an anomalous decline, which exposes the critical issue of quality variance in LLM-generated prompts. While expanding prompt volume provides richer contextual perspectives that help build more robust and generalizable medical concept representations, inaccurate or low-quality prompts conversely introduce noise interference that degrades model performance.

\begin{table}[htbp]
\centering
\caption{Biomed-DPT's classification accuracy (\%) across different $K$-shot settings with varying numbers of LLM prompts.}
\label{dif_prompt}
\begin{tabular}{lccccccc}
\hline
Prompts & K=0 & K=1 & K=2 & K=4 & K=8 & K=16 \\
\hline
10 & 50.04 & 55.14 & 57.80 & 62.42 & 67.03 & 69.51 \\
20 & 53.24 & 57.27 & 59.42 & 64.10 & 68.69 & 71.29 \\
30 & 55.67 & 58.42 & 60.83 & 65.62 & 70.26 & 72.94 \\
40 & 55.08 & 58.07 & 60.54 & 65.21 & 70.05 & 72.96 \\
50 & \textbf{56.02} & \textbf{59.03} & \textbf{61.27} & \textbf{66.12} & \textbf{70.76} & \textbf{73.51} \\
\hline
\end{tabular}
\end{table}

\subsubsection{Effect of Different CLIP-based Models}
This section compares four CLIP-based VLMs, including CLIP (ViT-B/16), PubMedCLIP (ViT-B/32), PMC-CLIP (RN50), and BiomedCLIP (ViT-B/16), to evaluate the impact of VLMs on Biomed-DPT.

Table \ref{avg_backbone} presents the few-shot classification performance of Biomed-DPT with different VLMs. As the number of samples increases, the accuracy of all VLMs improves. Biomed-DPT based on PubMedCLIP achieves a 39\% improvement in accuracy with 16-shot compared to zero-shot, fully demonstrating the effectiveness of Biomed-DPT. When using the 16-shot setting, Biomed-DPT with BiomedCLIP achieves a classification accuracy of 73.51\%, significantly outperforming VLMs. When the sample size increases from 8-shot to 16-shot, all VLMs maintain a stable performance improvement trend. 

These results not only confirm the superiority of BiomedCLIP as a domain-specific VLMs for biomedical applications but also demonstrate that Biomed-DPT, by effectively integrating the knowledge representation capabilities of specialized VLMs, can achieve significant performance improvements under few-shot conditions.

\begin{table}[htbp]
\centering
\caption{Comparison of classification accuracies (\%) of different CLIP-based backbone models in Biomed-DPT across various few-shot settings.}
\label{avg_backbone}
\begin{tabular}{lccccccc}
\hline
Method & K=0 & K=1 & K=2 & K=4 & K=8 & K=16 \\
\hline
CLIP~\cite{clip} & 24.44 & 43.37 & 45.78 & 51.32 & 53.67 & 61.52 \\
PubMedCLIP~\cite{pubmedclip} & 27.28  & 45.16 & 49.87 & 56.91 & 62.53 & 66.66 \\
PMC-CLIP~\cite{lin2023pmc} & 20.92  & 35.83 & 39.39 & 43.67 & 46.22 & 52.25 \\
BiomedCLIP~\cite{biomedclip} & \textbf{42.05} & \textbf{59.03} & \textbf{61.27} & \textbf{66.12} & \textbf{70.76} & \textbf{73.51} \\
\hline
\end{tabular}
\end{table}

\subsubsection{Visual Interpretability}
\begin{figure*}[htbp]
    \centering
    \setlength{\tabcolsep}{1pt} % 设置列间距为1pt
    \renewcommand{\arraystretch}{0} 
    \setlength{\extrarowheight}{1pt} % 添加额外的行间距
    \begin{tabular}{c*{6}{c}} 
        % --- 第1行 ---
        \raisebox{1.8\height}{\rotatebox[origin=c]{90}{\textbf{BTMRI}}}\hspace{5pt} &
        \includegraphics[width=0.15\textwidth,height=0.11\textheight]{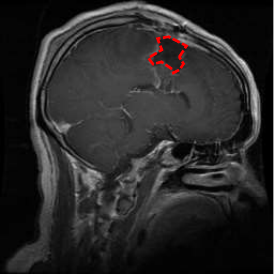} &
        \includegraphics[width=0.15\textwidth,height=0.11\textheight]{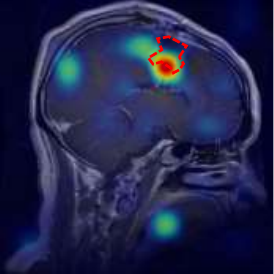} &
        \includegraphics[width=0.15\textwidth,height=0.11\textheight]{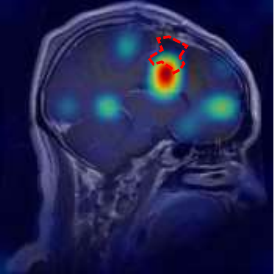} &
        \includegraphics[width=0.15\textwidth,height=0.11\textheight]{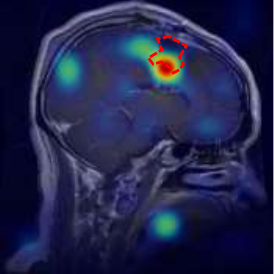} &
        \includegraphics[width=0.15\textwidth,height=0.11\textheight]{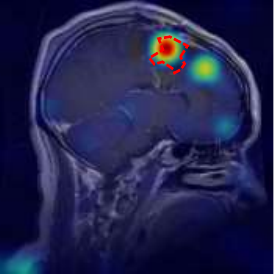} &
        \includegraphics[width=0.15\textwidth,height=0.11\textheight]{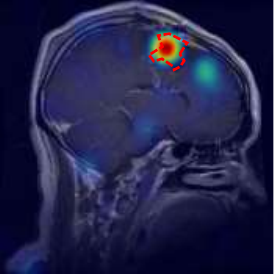} \\[2pt] % 添加行间距

        % --- 第2行 ---
        \raisebox{2.4\height}{\rotatebox[origin=c]{90}{\textbf{BUSI}}}\hspace{5pt} &
        \includegraphics[width=0.15\textwidth,height=0.11\textheight]{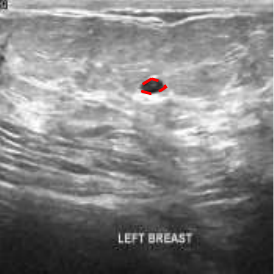} &
        \includegraphics[width=0.15\textwidth,height=0.11\textheight]{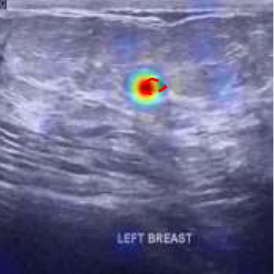} &
        \includegraphics[width=0.15\textwidth,height=0.11\textheight]{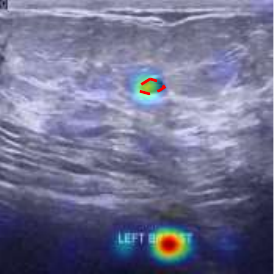} &
        \includegraphics[width=0.15\textwidth,height=0.11\textheight]{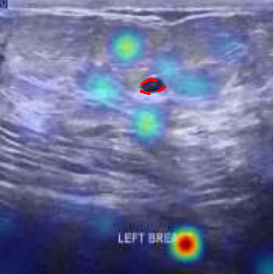} &
        \includegraphics[width=0.15\textwidth,height=0.11\textheight]{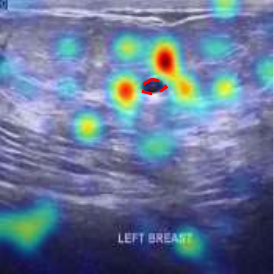} &
        \includegraphics[width=0.15\textwidth,height=0.11\textheight]{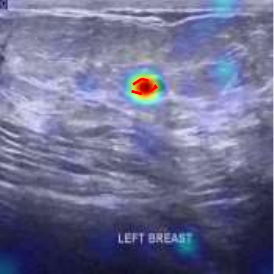} \\[2pt] % 添加行间距

        % --- 第3行 ---
        \raisebox{0.9\height}{\rotatebox[origin=c]{90}{\textbf{DermaMNIST}}}\hspace{5pt} &
        \includegraphics[width=0.15\textwidth,height=0.11\textheight]{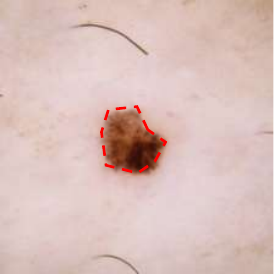} &
        \includegraphics[width=0.15\textwidth,height=0.11\textheight]{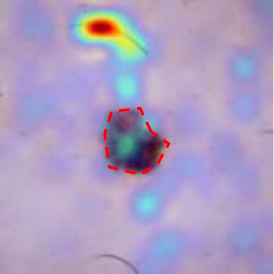} &
        \includegraphics[width=0.15\textwidth,height=0.11\textheight]{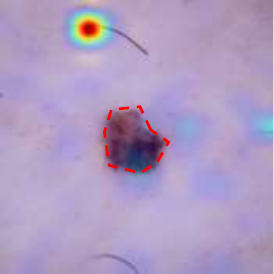} &
        \includegraphics[width=0.15\textwidth,height=0.11\textheight]{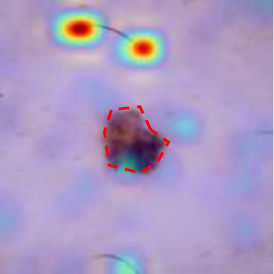} &
        \includegraphics[width=0.15\textwidth,height=0.11\textheight]{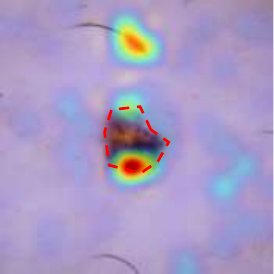} &
        \includegraphics[width=0.15\textwidth,height=0.11\textheight]{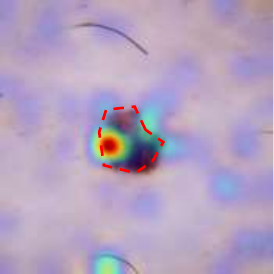} \\[2pt] % 添加行间距

        % --- 第4行 ---
        \raisebox{1.4\height}{\rotatebox[origin=c]{90}{\textbf{RETINA}}}\hspace{5pt} &
        \includegraphics[width=0.15\textwidth,height=0.11\textheight]{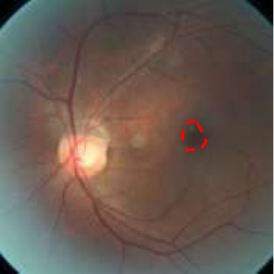} &
        \includegraphics[width=0.15\textwidth,height=0.11\textheight]{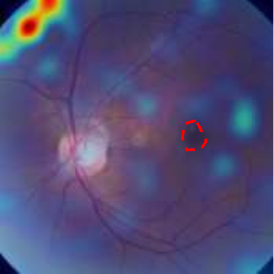} &
        \includegraphics[width=0.15\textwidth,height=0.11\textheight]{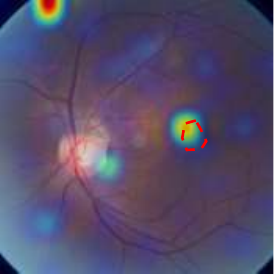} &
        \includegraphics[width=0.15\textwidth,height=0.11\textheight]{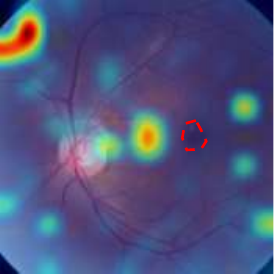} &
        \includegraphics[width=0.15\textwidth,height=0.11\textheight]{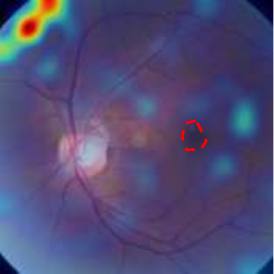} &
        \includegraphics[width=0.15\textwidth,height=0.11\textheight]{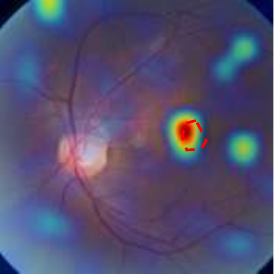} \\[5pt] % 添加行间距

        % --- 列标题（最后一行下方）---
        & \footnotesize (a) Image & \footnotesize (b) BiomedCLIP & \footnotesize (c) CoOp 
        & \footnotesize (d) BiomedCoOp & \footnotesize (e) w/o ZSP & \footnotesize (f) Biomed-DPT \\
    \end{tabular}
    \caption{Effect of various text prompt techniques on visual saliency maps. Columns (b)-(f) represent different prompt methods.}
    \label{fig:grid}
\end{figure*}

This section employs GradCAM~\cite{gradcam} to evaluate the impact of text prompts and the use of zero vector as a soft prompts on the visual saliency maps of biomedical images. Each column in Figure~\ref{fig:grid} represents a different prompt type: (a) Image, (b) BiomedCLIP, (c) CoOp, (d) BiomedCoOp, (e) Biomed-DPT w/o ZSP, and (f) Biomed-DPT.

BiomedCLIP (column b) exhibits notably dispersed attention features, failing to effectively focus on lesion regions in both the DermaMNIST and RETINA datasets, instead concentrating on non-diagnostic areas. In contrast, the CoOp method (column c) primarily attends to lesion regions, though for the DermaMNIST dataset, attention still localizes to non-diagnostic areas. For Biomed-DPT w/o ZSP (column e), attention is consistently directed toward lesion regions, but in the RETINA dataset, it remains concentrated on non-diagnostic areas. Meanwhile, Biomed-DPT (column f) demonstrates the best lesion localization capability across all biomedical imaging modalities, accurately highlighting clinically relevant regions.

In modalities where BiomedCLIP performs relatively well, such as MRI and ultrasound, Biomed-DPT shows minor improvements. However, in modalities where BiomedCLIP originally performs poorly, such as fundus photography and OCT, where most attention is scattered across non-diagnostic regions, Biomed-DPT significantly enhances performance by greatly reducing false positives and negatives. This precise localization improves interpretability, making it particularly valuable for biomedical applications where explainability is crucial. These observations further validate the effectiveness of our proposed method.

\section{Conclusion}
The pre-trained VLMs have demonstrated remarkable accuracy in natural image classification tasks. However, the high-dimensional characteristics of biomedical images and the dense professional terminology in clinical reports pose unique challenges for fine-tuning. Biomed-DPT in this study addresses these challenges through a knowledge-enhanced dual modality prompt tuning approach, making improvements in three key aspects: 1) In visual representation, it introduces the zero vector as a soft prompt to effectively suppress interference from non-diagnostic regions; 2) In text understanding, it designs specialized clinical prompt templates and leverages LLMs to incorporate external domain knowledge; 3) It employs knowledge distillation techniques to achieve efficient knowledge transfer in student models.

The effectiveness of Biomed-DPT is comprehensively evaluated using 11 open-source datasets, covering 10 lesion types across different organs and 9 imaging modalities. The comparative analysis included 9 fine-tuning methods, comprising 2 adapter-based fine-tuning, 2 linear probe-based fine-tuning, and 5 text prompt learning. The impact of different VLMs on the fine-tuning was also assessed. The results demonstrate that Biomed-DPT exhibits superior accuracy in classification tasks. Additionally, through visual analysis, we validate the effectiveness of zero vector and reveal the model's exceptional capability in lesion localization.

Although Biomed-DPT demonstrates strong performance in prompt tuning, it still faces several limitations. First, to further enhance its robustness and generalization capability, external domain knowledge generated by LLMs is required. This imposes strict demands on domain expertise, while LLMs do not always produce accurate and reasonable domain knowledge. Incorrect domain knowledge can significantly degrade model accuracy. Second, Biomed-DPT has not yet fully exploited the interactive potential between visual and text prmopts. Future work could explore synergistic optimization mechanisms for cross-modal prompts. Finally, challenges related to clinical deployment, including data privacy and ethical considerations, need to be investigated in future research.

In summary, Biomed-DPT combines the characteristics of biomedical images with BiomedCLIP to construct a classification system that balances generalization capability and task adaptability. Both fine-tuning experiments and Base-to-Novel generalization tests demonstrate the method's significant value in advancing the development of foundational models for biomedical applications. This work opens new avenues for developing more efficient and versatile AI solutions in healthcare, contributing positively to the advancement of smart medicine.

\bibliographystyle{unsrt}
\bibliography{references}

\appendices

\newpage

\section*{Appendix and the Use of Supplemental Files}
\renewcommand{\thefigure}{\thesection\arabic{figure}}
\setcounter{figure}{0}
\renewcommand{\thetable}{\thesection\arabic{table}}
\setcounter{table}{0}
\subsection{Additional Few-shot Results}
\begin{figure}[!t]  
    \centering
    \includegraphics[width=0.7\linewidth]{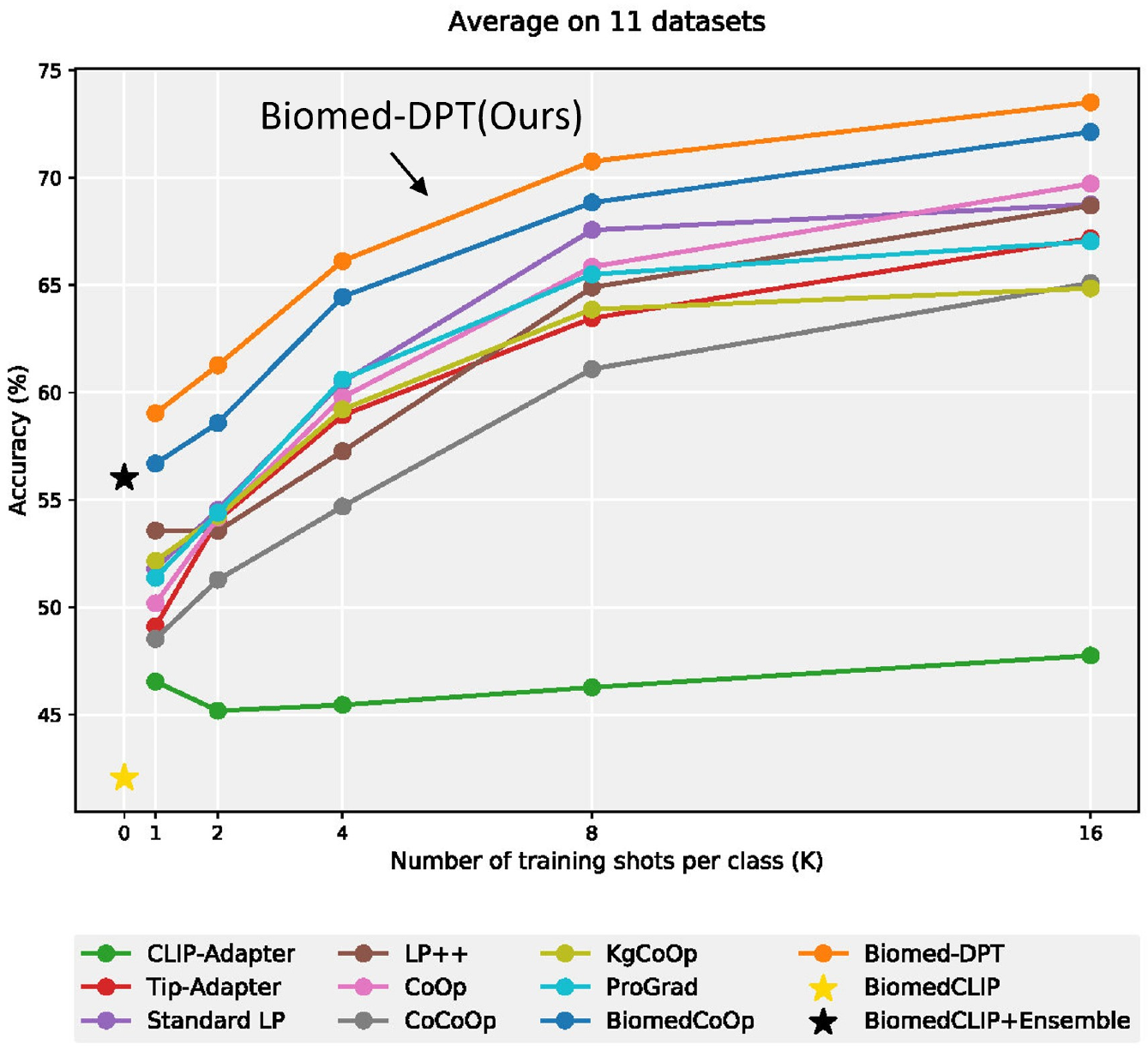}
    \caption{Average classification accuracy (\%) of various fewshot adaptation methods across different numbers of training shots per class.}
    \label{fig:ax_avg_method}
\end{figure}

Figure ~\ref{fig:ax_avg_method} illustrates the performance of Biomed-DPT and various fewshot adaptation methods across different few-shot settings ($K=1, 2, 4, 8, 16$), highlighting Biomed-DPT’s robustness in low-data regimes. Overall, Biomed-DPT not only matches but frequently surpasses SOTA PEFT methods across diverse benchmarks.

\subsection{Effect of Context}
\begin{figure*}[!t]  
    \centering
    \includegraphics[width=0.87\linewidth]{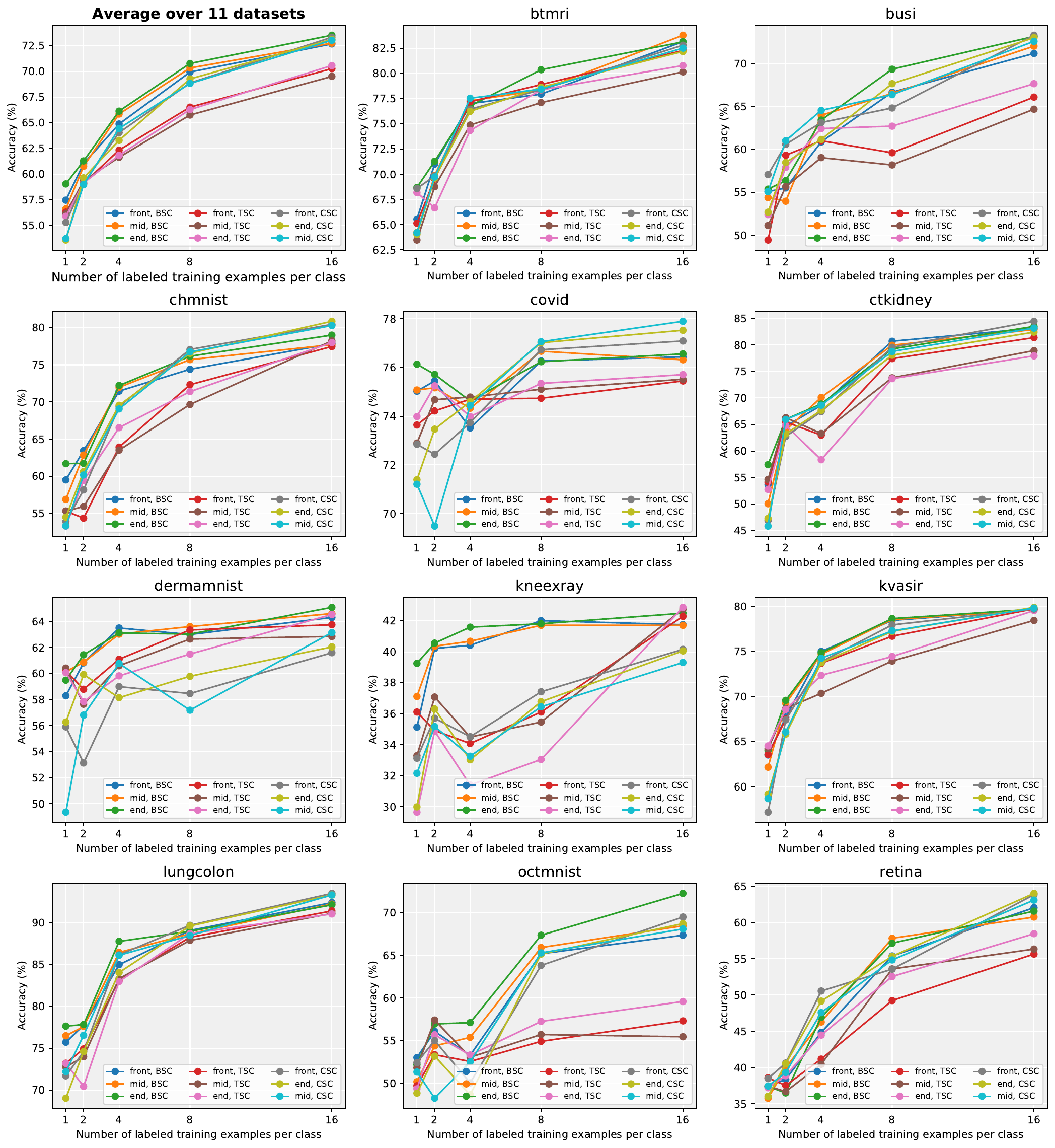}
    \caption{Comparison with differnet context on the 11 datasets. ``front'' or ``mid'' or ``end''  means putting the class token in the front or middle or end. TSC means task-specific context. BSC means biomedical-specific context. CSC means class-specific context. }
    \label{fig:ax_main_results}
\end{figure*}

This experiment aims to investigate the impact of different context types and their insertion positions on model accuracy. We compared the model performance when using biomedical-specific context (BSC), class-specific context (CSC), and task-specific context (TSC) as contextual cues, and further analyzed how different insertion positions (front, mid, end) within the input sequence affect performance. Figure~\ref{fig:ax_main_results} shows that BSC achieves the highest model accuracy, indicating that biomedical information can more effectively guide the model in understanding task requirements and improving predictive performance. CSC performs second-best, suggesting that class-specific context provides more informative class-related cues compared to task-specific context. Regarding insertion position, placing the context at the end yields the most significant improvement, followed by mid, while front positioning performs the worst. This demonstrates that the optimal insertion position for context is at the end of the sequence.

\subsection{Learnable Context Interpretability}
This study analyzes the context tokens learned from different biomedical datasets and their semantic neighbors, thoroughly investigating the mapping relationships between these abstract representations and visual features in medical imaging. As shown in Table~\ref{tabl:ax_cntx_words}, we not only quantify the nearest neighbor words (with Euclidean distances) of each context token in the embedding space but also observe that the model accurately identifies modality-specific descriptors for different biomedical imaging types. Specifically, domain-distinctive terms such as ``\texttt{endoscopic}'' in the Kvasir dataset, ``\texttt{mr}'' in the BTMRI dataset, and ``\texttt{histopathological}'' in the LC25000 dataset are effectively captured. These findings demonstrate that the model can autonomously establish semantic associations between multimodal medical images and their corresponding professional descriptions, revealing its deep contextual understanding of cross-modal biomedical data.

\begin{table}[!t]
\centering
\small
\caption{Hyperparameter settings for $\lambda_1$  and $\lambda_2$ across different datasets}
\label{tab:hyperparams}
\resizebox{0.4\linewidth}{!}{
\begin{tabular}{lc|cc}
\toprule
Dataset & Benchmark & $\lambda_1$ & $\lambda_2$ \\
\midrule
\multirow{2}{*}{\centering BTMRI}  & Few-shot & 12.50 & 0.25 \\
& Base-to-Novel & 25.00 & 0.50 \\
\midrule
BUSI &Few-shot & 18.75 & 1.00 \\
& Base-to-Novel & - & - \\
\midrule
CHMNIST & Few-shot & 12.50 & 0.50 \\
& Base-to-Novel & 23.00 & 0.90 \\
\midrule
COVID-QU-Ex& Few-shot & 65.00 & 2.50 \\
& Base-to-Novel & 18.75 & 0.50 \\
\midrule
CTKIDNEY& Few-shot & 18.75 & 0.25 \\
& Base-to-Novel & 25.00 & 1.00 \\
\midrule
DermaMNIST& Few-shot & 25.00 & 20.00 \\
& Base-to-Novel & 11.50 & 5.00 \\
\midrule
KneeXray & Few-shot & 25.00 & 10.00 \\
& Base-to-Novel & 23.00 & 4.50 \\
\midrule
Kvasir& Few-shot & 25.00 & 0.75 \\
& Base-to-Novel & 10.00 & 0.50 \\
\midrule
LC25000 & Few-shot & 12.50 & 0.25 \\
& Base-to-Novel & 9.50 & 0.30 \\
\midrule
OCTMNIST & Few-shot & 12.50 & 0.25 \\
& Base-to-Novel & 12.50 & 0.50 \\
\midrule
RETINA & Few-shot & 10.50 & 0.25 \\
& Base-to-Novel & 22.00 & 0.40 \\
\bottomrule
\end{tabular}
}
\end{table}

\begin{table*}[htbp]
\centering
\caption{The nearest words for each of the 5 context vectors learned by Biomed-DPT, with their distances to the corresponding context tokens shown in parentheses.}
\label{tabl:ax_cntx_words}
\resizebox{\linewidth}{!}{
\begin{tabular}{cccccc}
\hline
Dataset & Token \#1 & Token \#2 & Token \#3 & Token \#4 & Token \#5 \\
\hline
BTMRI & a (1.1756) & \textbf{mr (2.1553)} & photo (1.7428) & of (1.1224) & a (1.4610) \\
BUSI & a (0.8401) & \textbf{ultrasound (1.3669)} & photo (1.1629) & of (0.8260) & a (1.2984) \\
CHMNIST & grain (2.5166) & \textbf{histopathological (3.1773)} & are (2.4925) & in (2.2326) & when (2.7661) \\
COVID-QU-Ex & a (0.8142) & \textbf{chest (1.3873)} & x (1.1960) & - (0.8581) & ray (1.2888) \\
CTKidney & a (0.8580) & ct (1.8385) & \textbf{photo (2.0514)} & of (0.8325) & a (1.1857) \\
DermaMNIST & superc (2.5139) & \textbf{rash (4.5743)} & concern (4.0883) & photo (3.0455) & recognizes (2.5360) \\
KneeXray & a (1.4196) & \textbf{frontal (1.8284) }& x (1.5142) & - (1.2100) & ray (1.8029) \\
Kvasir & a (1.9554) & \textbf{endoscopic (2.6161)} & photo (1.9878) & of (1.6157) & a (1.8610) \\
LC25000 & a (1.2181) & \textbf{histopathological (2.2258)} & photo (1.6093) & of (1.5616) & a (1.6273) \\
OCTMNIST & a (1.3193) & \textbf{oct (2.1375)} & photo (1.2527) & of (1.2719) & a (1.1053) \\
RETINA & \textbf{\#\#atology (2.1274)} & photo (1.5691) & of (1.9162) & infarcts (1.9107) & complained (1.8346) \\
\hline
\end{tabular}
}
\end{table*}

\subsection{Additional Hyperparameters}
Table ~\ref{tab:hyperparams} outlines the selected hyperparameters ($\lambda_1$, $\lambda_2$) used across various datasets for Biomed-DPT’s few-shot and Base-to-Novel benchmarks. These parameters were optimized to balance classification accuracy and model adaptability, with $\lambda_1$ and $\lambda_2$ controlling the weight of the regularization and distillation losses.
\end{document}